\documentclass{article}

\PassOptionsToPackage{numbers,compress}{natbib}

\usepackage[dblblindworkshop, final]{neurips_2026}

\workshoptitle{Continual Learning in the Era of Foundation Models and Embodied Agents}

\usepackage[utf8]{inputenc} 
\usepackage[T1]{fontenc}    
\usepackage{hyperref}       
\usepackage{url}            
\usepackage{booktabs}       
\usepackage{amsfonts}       
\usepackage{nicefrac}       
\usepackage{microtype}      
\usepackage{xcolor}         
\usepackage{array}
\usepackage{tabularx}
\usepackage{algorithm}
\usepackage{algorithmic}
\usepackage{multirow}
\usepackage{makecell}
\usepackage{booktabs}
\usepackage{multirow}
\usepackage{adjustbox}
\usepackage{xcolor}
\usepackage{pifont}
\usepackage{colortbl}

\usepackage{booktabs}
\usepackage{amsmath}
\usepackage{amssymb}
\usepackage{mathtools}
\usepackage{amsthm}
\usepackage{tabularx}
\theoremstyle{plain}

\theoremstyle{definition}

\theoremstyle{remark}

\usepackage[table]{xcolor}
\usepackage{enumitem}
\definecolor{stronggreen}{HTML}{A9D6B5}
\definecolor{midgreen}{HTML}{CDE8D3}
\definecolor{lightgreen}{HTML}{EAF5EC}
\definecolor{lightred}{HTML}{F4CCCC}

\title{Do Vision CIL Methods Transfer to Protein? A Diagnostic Benchmark of Frozen Protein Encoders}

\author{%
  Ziyuan Gao \\
  Department of Computer Science and Technology\\
  University of Cambridge\\
  \texttt{zg344@cam.ac.uk} \\
}

\begin{document}

\maketitle

\begin{abstract}
Protein function annotation operates over an expanding label space, making class-incremental learning (CIL) a natural framework for incorporating newly discovered functions without full retraining. Yet existing CIL methods have been developed largely in vision and have not been systematically evaluated over frozen protein representations. We conduct a controlled benchmark of nine representative CIL methods across five sequence-, structure-, and sequence--structure-based protein encoders and two enzyme function datasets at different scales, while keeping the encoders frozen and evaluation settings consistent across representations. We find that no method transfers reliably across encoders, and that these failures persist across alternative phase constructions. Diagnostic analyses reveal that this instability arises from weak gradient-based stability signals, substantial heterogeneity in embedding geometry, and covariance assumptions whose appropriate capacity changes with both dataset scale and encoder. These findings suggest that the representation itself should be treated as a central component of continual-learning design rather than a passive feature extractor. More generally, reliable protein CIL will require encoder-aware methods that adapt their statistical assumptions and capacity to the geometry of the representation on which they operate. Code is available at https://github.com/zygao930/ProteinCIL.
\end{abstract}

\section{Introduction}

Protein function annotation operates over a label space that continues to grow. Taking Enzyme Commission (EC) numbers as an example, the Swiss-Prot database expanded from roughly 4{,}700 EC numbers in 2018 to over 5{,}000 in 2024 \citep{uniprot2023uniprot}. A deployed annotation model should therefore absorb newly discovered enzyme classes without retraining on the full historical database whenever the label space changes \citep{masana2022class, sledzieski2025copep}. Class-incremental learning (CIL) provides a natural framework for this setting: a model learns new classes sequentially while preserving discriminative ability on those seen before \citep{masana2022class, rebuffi2017icarl, kirkpatrick2017overcoming}. At the same time, protein annotation increasingly relies on pretrained encoders that provide reusable representations without task-specific backbone training \citep{lin2023evolutionary, zhang2023protein}. This raises a question that has not been systematically examined: do CIL methods developed largely for computer vision transfer to frozen protein representations?

Two lines of work meet at this question. On one side, protein representation learning has shifted from task-specific supervised models toward general-purpose pretrained encoders. Sequence-based protein language models such as TAPE \citep{rao2019tape}, ESM \citep{rives2021biological}, ESM2 \citep{lin2023evolutionary}, and ProtTrans \citep{elnaggar2022prottrans} capture structural and functional signal from sequence, while structure-aware models such as GearNet \citep{zhang2023protein}, SaProt \citep{su2024saprot}, and PST \citep{chen2024pst} incorporate geometric information. These encoders support enzyme function and EC-number prediction \citep{yu2023enzyme, hermosilla2021intrinsic, yang2024cleancontact, wang2025dpfunc, schmirler2024finetuning}, and their embedding spaces organize functionally related proteins \citep{wang2025splm, holm2023tree}. Yet this work largely assumes a fixed label space \citep{rao2019tape, yu2023enzyme}. On the other side, class-incremental learning studies sequential class arrival \citep{masana2022class}. Classical methods address forgetting through regularization, replay, exemplar memory, or distillation, including EWC \citep{kirkpatrick2017overcoming}, iCaRL \citep{rebuffi2017icarl}, A-GEM \citep{chaudhry2019tiny}, DER++ \citep{buzzega2020dark}, and RER \citep{lin2024rer}. 
Recent approaches increasingly build on pretrained backbones \citep{11492684,pmlr-v306-gao26s} through prompt tuning \citep{wang2022dualprompt, smith2023coda}, classifier alignment \citep{zhang2023slca}, random projection \citep{mcdonnell2023ranpac}, subspace ensembles \citep{wang2024ease}, and model surgery \citep{sun2025mos}.
 Prototype-based methods provide a complementary route from feature statistics \citep{mensink2013distance}, with FeCAM \citep{goswami2024fecam} using covariance-aware Mahalanobis classification with shrinkage estimation \citep{ledoit2004well, chen2010shrinkage}. These methods have been developed and validated largely on vision benchmarks, and whether their assumptions transfer to frozen protein representations remains untested.

\begin{figure}[t]
\centering
\includegraphics[width=0.8\columnwidth]{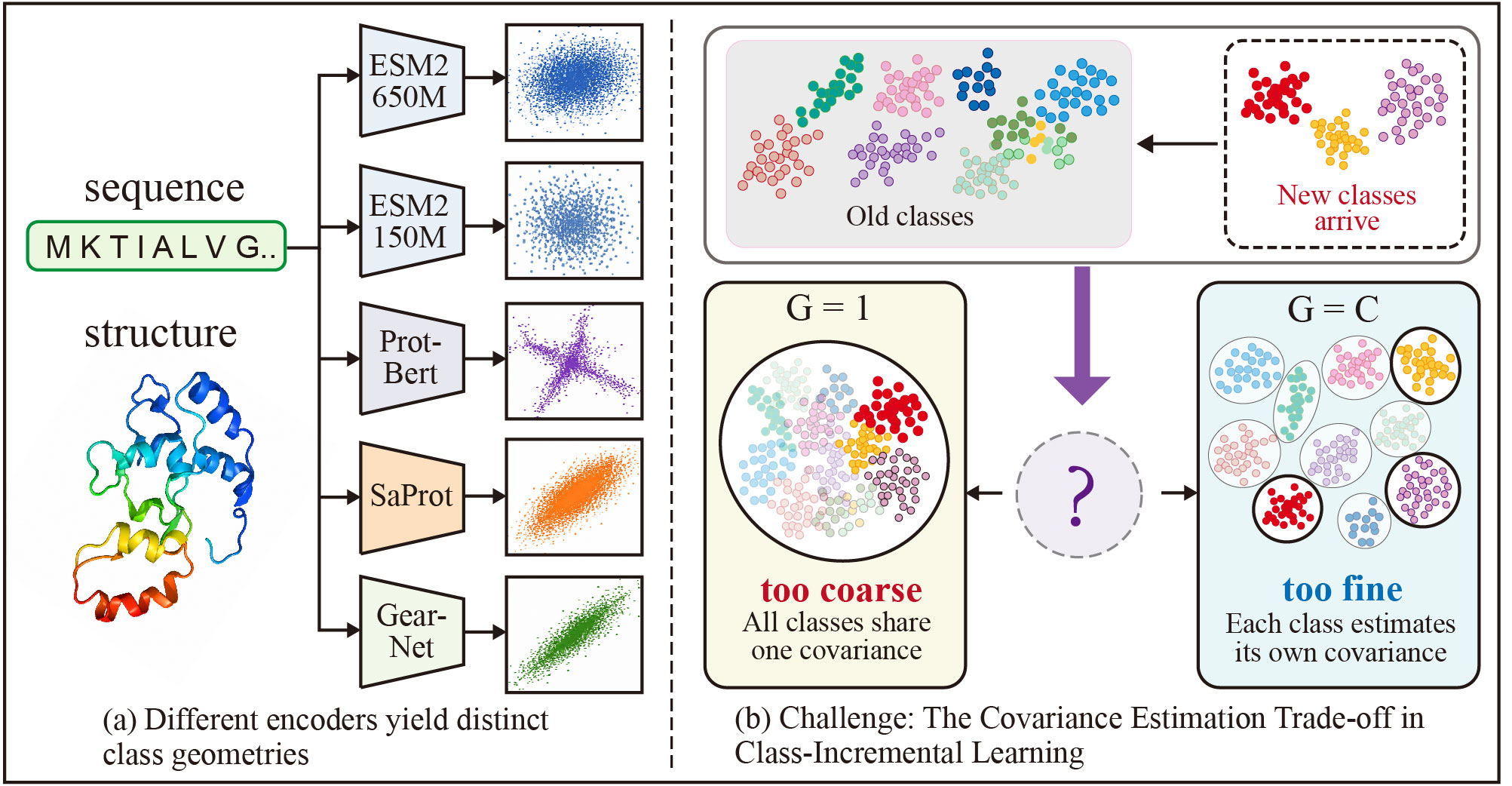}
\caption{Motivation for frozen protein CIL. Frozen protein encoders induce distinct class geometries, creating a covariance-capacity trade-off as new classes arrive.}
\label{fig:motivation}
\end{figure}

To answer this, we conduct a controlled benchmark of nine representative CIL methods, with Naive fine-tuning and Joint training as lower and offline reference points, across five frozen protein encoders and two enzyme datasets at different scales. The encoders span sequence language models, a structure-based graph network, and a sequence--structure fusion model, and differ substantially in embedding dimension and geometry, as illustrated in Fig.~\ref{fig:motivation}. We use four phase-construction protocols, including EC-aligned and alternative class groupings, to test whether conclusions depend on the incremental stream. Each method uses one fixed hyperparameter configuration across encoders and datasets, directly testing the plug-and-play transfer setting rather than encoder-specific tuning. The result is consistent across protocols: no CIL method is reliably strong across representations, and the dominant variation is encoder-dependent rather than phase-dependent. Diagnostic analyses trace this instability to three recurring causes: weak gradient-based stability signals when only the classifier head is trainable, large differences in encoder-induced geometry, and covariance capacity requirements that change jointly with dataset scale and representation geometry.

Our contributions are threefold. First, we establish the first systematic benchmark of CIL on frozen protein representations, spanning nine CIL methods, five encoders, two enzyme datasets, and four phase construction protocols. Across this benchmark, no existing method transfers reliably across encoders and dataset scales. Second, we show that this instability is primarily representation dependent rather than phase dependent. Method rankings vary substantially across encoders, while the performance patterns remain stable under alternative phase constructions. We further trace this instability to weak gradient based stability signals, variation in embedding geometry, and covariance capacity requirements that change with both dataset scale and encoder geometry. Third, we show that these failures reflect a mismatch between the fixed assumptions of existing CIL methods and the representations on which they operate. This points toward encoder aware protein CIL, where statistical capacity adapts to representation geometry rather than being fixed in advance.

\section{Method}
\label{sec:framework}

We formalize class-incremental learning over frozen protein encoders, then describe the evaluation protocol and the taxonomy of methods our benchmark spans. Concrete datasets, encoder extraction details, and hyperparameters are deferred to the experimental setup.

\paragraph{Problem formulation.}
We consider class-incremental learning over a sequence of $T$ phases $\mathcal{P}=\{P_0,P_1,\ldots,P_{T-1}\}$, where each phase $P_t$ introduces a disjoint set of new classes. A frozen pretrained protein encoder $f_\theta$ maps each protein to a $d$-dimensional embedding
\[
x=f_\theta(\text{protein})\in\mathbb{R}^d,
\]
and its parameters $\theta$ are never updated. After observing phase $P_t$, the learner must classify any query protein over all classes encountered so far,
\[
C_{\leq t}=\bigcup_{i=0}^{t}P_i,
\]
using the frozen protein embeddings together with the method-specific components learned or maintained during the incremental sequence. This setting separates continual-learning behavior from backbone adaptation~\citep{zhou2024continual, zhou2023revisiting}, allowing different incremental-learning strategies to be compared under a fixed representation space.

\paragraph{Benchmark workflow.}
We evaluate all methods within a single controlled pipeline. Each protein is first encoded by a frozen backbone, and the resulting embeddings are cached for all subsequent experiments. We include five frozen protein encoders with different geometric properties: sequence-based models ESM2-650M ($d{=}1280$), ESM2-150M ($d{=}640$)~\citep{lin2023evolutionary}, and ProtBert ($d{=}1024$)~\citep{elnaggar2022prottrans}; structure-based GearNet-Edge ($d{=}3072$)~\citep{zhang2023protein}; and the sequence-structure fusion model SaProt ($d{=}1280$)~\citep{su2024saprot}. The encoders remain frozen throughout. Method-specific hyperparameters are fixed across all encoder-dataset configurations, with no setting-specific tuning.

Classes are introduced under four phase construction protocols that vary in their alignment with the Enzyme Commission (EC) hierarchy~\citep{webb1992enzyme}. \textbf{EC-first} assigns one phase to each top-level EC category; \textbf{Random} shuffles classes across phases while preserving the EC-first phase sizes; \textbf{Finer-EC} groups classes using a finer EC prefix and greedily packs these groups into phases; and \textbf{Mixed} combines classes from different top-level EC groups within each phase. Phase sizes are kept comparable across protocols so that performance differences are not driven simply by the number of classes introduced at each step. For each random seed, phase order is shuffled once, and all methods evaluated under that seed observe exactly the same incremental stream.

\paragraph{Methods and evaluation.}
Nine CIL methods are evaluated under this common protocol, with Naive fine-tuning and Joint training serving as incremental lower and offline upper reference points, respectively. Regularization-based methods are represented by EWC~\citep{kirkpatrick2017overcoming}. Replay-based methods include Replay~\citep{chaudhry2019tiny}, iCaRL~\citep{rebuffi2017icarl}, DER++~\citep{buzzega2020dark}, and RER~\citep{lin2024rer}, which retain examples from previous phases and differ in how stored information is reused or distilled. Prototype- and covariance-based methods include FeCAM~\citep{goswami2024fecam} and FeTrIL~\citep{petit2023fetril}, which update class representations using feature statistics. Projection- and subspace-based methods include RanPAC~\citep{mcdonnell2023ranpac} and EASE~\citep{wang2024ease}, which construct classifiers in transformed embedding spaces.

After each phase, every method is evaluated on all classes observed so far. This produces a lower triangular accuracy matrix
\[
R \in \mathbb{R}^{T \times T},
\]
where $R[i][j]$ denotes the classification accuracy on phase $P_j$ after training through phase $P_i$, with $j \leq i$. From $R$, we report three standard continual learning metrics~\citep{masana2022class}. Average accuracy measures final retained performance~\citep{lopez2017gradient}, forgetting measures degradation from the best historical performance~\citep{chaudhry2018riemannian}, and plasticity measures new-class acquisition at each phase:
\begin{subequations}
\label{eq:cil_metrics}
\begin{align}
\mathrm{AvgAcc} &= \frac{1}{T}\sum_{j=0}^{T-1} R[T{-}1][j],
\label{eq:avg_acc}\\
\mathrm{Fgt} &= \frac{1}{T-1}\sum_{j=0}^{T-2}
\left(\max_{i\in[j,T-2]} R[i][j] - R[T{-}1][j]\right),
\label{eq:forgetting}\\
\mathrm{Plast} &= \frac{1}{T}\sum_{i=0}^{T-1} R[i][i].
\label{eq:plasticity}
\end{align}
\end{subequations}
All reported results are averaged over random seeds.

\section{Results}
\label{sec:Results}


\subsection{Overall results.}

Table~\ref{tab:main} reveals that method rankings are unstable across
encoders and datasets: the strongest baseline on one encoder is often
mediocre on another, and no method dominates all five encoders on either
dataset. Regularization (EWC) provides inconsistent gains over the naive
lower bound, consistent with the limited gradient signal available when
the encoder is frozen. Replay-based methods retain high new-class
plasticity but still incur substantial forgetting, particularly on
EC~Number. The two prototype methods fail in opposite directions:
FeTrIL retains high plasticity yet forgets severely as classes
accumulate, while FeCAM preserves old classes effectively but degrades
sharply on EC~Number, where its plasticity falls near the bottom of the
table. RanPAC is strong on most sequence and sequence--structure
encoders but deteriorates sharply on GearNet-Edge. Overall, transferring
vision-based CIL recipes to protein encoders yields no uniformly reliable
winner, motivating a diagnosis of the representation-dependent failure
modes.

\begin{table*}[h]
\centering
\caption{Benchmark results across two datasets and five encoders. Encoder-specific results are reported as mean $\pm$ std over 3 seeds. Best per column in \textbf{bold}, second \underline{underlined} (Naive and Joint excluded from ranking). Category abbreviations: Reg.\ = Regularization, Rep.\ = Replay, Sub.\ = Subspace, Proj.\ = Projection, Proto.\ = Prototype.}
\label{tab:main}
\resizebox{\textwidth}{!}{%
\renewcommand{\arraystretch}{1}
\begin{tabular}{ll | ccccc | ccccc}
\toprule
& & \multicolumn{5}{c}{\textbf{Enzyme Reaction (348 classes)}} & \multicolumn{5}{c}{\textbf{EC Number (4788 classes)}} \\
\cmidrule(lr){3-7} \cmidrule(lr){8-12}
& & ESM2 & ESM2 & GearNet & & & ESM2 & ESM2 & GearNet & & \\
\textbf{Cat.} & \textbf{Method} & 650M & 150M & Edge & SaProt & ProtBert & 650M & 150M & Edge & SaProt & ProtBert \\
\midrule

\multicolumn{12}{c}{\textbf{\textit{Average Incremental Accuracy (\%) $\uparrow$}}}\\
\midrule
\textit{LB} & Naive
& 13.7\tiny{$\pm$0.5} & 12.6\tiny{$\pm$1.5} & 7.4\tiny{$\pm$1.4} & 13.4\tiny{$\pm$0.7} & 10.7\tiny{$\pm$0.5}
& 25.4\tiny{$\pm$1.6} & 23.5\tiny{$\pm$1.4} & 8.6\tiny{$\pm$0.3} & 20.7\tiny{$\pm$0.8} & 21.4\tiny{$\pm$0.2} \\

\textit{Joint} & Joint MLP
& 73.9\tiny{$\pm$0.8} & 69.1\tiny{$\pm$0.8} & 52.1\tiny{$\pm$1.1} & 71.2\tiny{$\pm$0.4} & 65.5\tiny{$\pm$0.3}
& 99.5\tiny{$\pm$0.2} & 99.5\tiny{$\pm$0.1} & 68.4\tiny{$\pm$0.7} & 99.4\tiny{$\pm$0.0} & 99.5\tiny{$\pm$0.0} \\
\cmidrule{1-12}

\textit{Reg.} & EWC
& 21.8\tiny{$\pm$4.0} & 17.7\tiny{$\pm$4.1} & 6.9\tiny{$\pm$1.1} & 20.8\tiny{$\pm$4.0} & 15.0\tiny{$\pm$1.8}
& 45.2\tiny{$\pm$5.4} & 40.2\tiny{$\pm$2.9} & 8.7\tiny{$\pm$1.2} & 32.6\tiny{$\pm$1.8} & 40.2\tiny{$\pm$1.3} \\
\cmidrule{1-12}

\multirow{4}{*}{\textit{Rep.}}
& Replay
& 57.2\tiny{$\pm$2.0} & 52.5\tiny{$\pm$1.5} & 34.5\tiny{$\pm$1.9} & 53.9\tiny{$\pm$0.6} & 44.9\tiny{$\pm$1.9}
& 58.2\tiny{$\pm$2.6} & 52.8\tiny{$\pm$2.2} & 25.7\tiny{$\pm$0.4} & 51.0\tiny{$\pm$1.5} & 51.5\tiny{$\pm$1.6} \\

& iCaRL
& 62.5\tiny{$\pm$0.6} & 56.4\tiny{$\pm$0.8} & \underline{36.9}\tiny{$\pm$0.8} & 60.1\tiny{$\pm$0.8} & 51.2\tiny{$\pm$0.4}
& 59.5\tiny{$\pm$1.7} & 52.9\tiny{$\pm$2.1} & 24.0\tiny{$\pm$1.9} & 50.6\tiny{$\pm$1.0} & 52.2\tiny{$\pm$0.9} \\

& DER++
& 66.8\tiny{$\pm$3.1} & 61.2\tiny{$\pm$5.2} & 33.1\tiny{$\pm$2.7} & 62.0\tiny{$\pm$1.8} & 54.7\tiny{$\pm$5.0}
& \underline{64.5}\tiny{$\pm$5.1} & \underline{59.4}\tiny{$\pm$5.7} & 25.9\tiny{$\pm$2.9} & 57.9\tiny{$\pm$6.7} & \underline{56.3}\tiny{$\pm$6.5} \\

& RER
& 56.4\tiny{$\pm$1.4} & 52.3\tiny{$\pm$2.1} & 34.6\tiny{$\pm$1.6} & 54.6\tiny{$\pm$1.0} & 45.8\tiny{$\pm$1.8}
& 58.2\tiny{$\pm$2.5} & 52.7\tiny{$\pm$2.3} & 25.8\tiny{$\pm$0.7} & 50.9\tiny{$\pm$1.5} & 51.4\tiny{$\pm$1.6} \\
\cmidrule{1-12}

\textit{Sub.} & EASE
& 55.4\tiny{$\pm$0.8} & 49.9\tiny{$\pm$0.8} & 29.4\tiny{$\pm$1.0} & 53.0\tiny{$\pm$0.9} & 43.7\tiny{$\pm$0.8}
& 60.0\tiny{$\pm$0.6} & 58.6\tiny{$\pm$0.3} & \textbf{44.7}\tiny{$\pm$0.2} & \underline{59.1}\tiny{$\pm$0.3} & 53.7\tiny{$\pm$0.9} \\
\cmidrule{1-12}

\textit{Proj.} & RanPAC
& \underline{74.3}\tiny{$\pm$0.0} & \textbf{70.5}\tiny{$\pm$0.0} & 21.8\tiny{$\pm$0.0} & \textbf{73.5}\tiny{$\pm$0.0} & \underline{65.8}\tiny{$\pm$0.0}
& \textbf{82.2}\tiny{$\pm$0.0} & \textbf{79.9}\tiny{$\pm$0.0} & 26.1\tiny{$\pm$0.0} & \textbf{84.7}\tiny{$\pm$0.0} & \textbf{79.0}\tiny{$\pm$0.0} \\
\cmidrule{1-12}

\multirow{2}{*}{\textit{Proto.}}
& FeTrIL
& 64.1\tiny{$\pm$6.6} & 46.9\tiny{$\pm$6.4} & 31.0\tiny{$\pm$4.5} & 47.6\tiny{$\pm$5.6} & 43.0\tiny{$\pm$4.9}
& 47.6\tiny{$\pm$6.9} & 46.5\tiny{$\pm$6.3} & 23.8\tiny{$\pm$4.2} & 46.0\tiny{$\pm$6.2} & 46.3\tiny{$\pm$6.5} \\

& FeCAM
& \textbf{75.0}\tiny{$\pm$0.0} & \underline{67.1}\tiny{$\pm$0.0} & \textbf{43.4}\tiny{$\pm$0.0} & \underline{72.5}\tiny{$\pm$0.0} & \textbf{69.6}\tiny{$\pm$0.0}
& 55.8\tiny{$\pm$0.0} & 51.7\tiny{$\pm$0.0} & \underline{37.7}\tiny{$\pm$0.0} & 47.7\tiny{$\pm$0.0} & 30.2\tiny{$\pm$0.0} \\

\midrule
\multicolumn{12}{c}{\textbf{\textit{Forgetting (\%) $\downarrow$}}}\\
\midrule

\textit{LB} & Naive
& 76.3\tiny{$\pm$1.2} & 73.3\tiny{$\pm$1.5} & 54.9\tiny{$\pm$3.0} & 73.3\tiny{$\pm$0.9} & 68.9\tiny{$\pm$1.0}
& 85.1\tiny{$\pm$2.3} & 86.3\tiny{$\pm$2.0} & 52.6\tiny{$\pm$5.7} & 87.6\tiny{$\pm$1.5} & 85.6\tiny{$\pm$0.7} \\
\cmidrule{1-12}

\textit{Reg.} & EWC
& 66.5\tiny{$\pm$4.6} & 66.4\tiny{$\pm$3.9} & 53.9\tiny{$\pm$2.4} & 65.2\tiny{$\pm$4.7} & 62.5\tiny{$\pm$1.8}
& 62.0\tiny{$\pm$6.5} & 66.7\tiny{$\pm$3.2} & 52.2\tiny{$\pm$4.5} & 73.7\tiny{$\pm$1.4} & 63.7\tiny{$\pm$2.2} \\
\cmidrule{1-12}

\multirow{4}{*}{\textit{Rep.}}
& Replay
& 23.0\tiny{$\pm$2.5} & 24.7\tiny{$\pm$2.1} & 24.2\tiny{$\pm$1.3} & 25.3\tiny{$\pm$0.2} & 28.0\tiny{$\pm$1.2}
& 47.5\tiny{$\pm$3.4} & 52.8\tiny{$\pm$3.7} & 36.1\tiny{$\pm$5.0} & 53.6\tiny{$\pm$2.8} & 51.7\tiny{$\pm$2.8} \\

& iCaRL
& 16.3\tiny{$\pm$1.4} & 20.1\tiny{$\pm$2.3} & 21.0\tiny{$\pm$1.9} & 17.4\tiny{$\pm$1.5} & 19.8\tiny{$\pm$1.3}
& 46.5\tiny{$\pm$2.3} & 53.6\tiny{$\pm$2.4} & 36.5\tiny{$\pm$6.6} & 54.9\tiny{$\pm$1.8} & 52.3\tiny{$\pm$1.3} \\

& DER++
& 12.7\tiny{$\pm$1.0} & 15.3\tiny{$\pm$3.1} & 23.2\tiny{$\pm$1.9} & 15.3\tiny{$\pm$1.1} & 16.0\tiny{$\pm$1.1}
& 39.7\tiny{$\pm$5.5} & 44.4\tiny{$\pm$5.8} & 26.4\tiny{$\pm$4.5} & 44.7\tiny{$\pm$6.7} & 45.3\tiny{$\pm$5.2} \\

& RER
& 24.5\tiny{$\pm$1.9} & 25.7\tiny{$\pm$2.9} & 22.6\tiny{$\pm$0.6} & 24.4\tiny{$\pm$1.0} & 27.2\tiny{$\pm$0.8}
& 47.5\tiny{$\pm$3.3} & 52.9\tiny{$\pm$3.7} & 35.5\tiny{$\pm$5.1} & 53.7\tiny{$\pm$2.8} & 51.7\tiny{$\pm$2.8} \\
\cmidrule{1-12}

\textit{Sub.} & EASE
& 11.0\tiny{$\pm$2.0} & 9.9\tiny{$\pm$3.4} & 12.7\tiny{$\pm$2.0} & 11.6\tiny{$\pm$1.7} & 9.8\tiny{$\pm$1.7}
& 8.0\tiny{$\pm$0.7} & \underline{6.7}\tiny{$\pm$0.3} & 5.8\tiny{$\pm$1.0} & 6.6\tiny{$\pm$1.2} & 9.1\tiny{$\pm$1.0} \\
\cmidrule{1-12}

\textit{Proj.} & RanPAC
& \underline{6.8}\tiny{$\pm$2.1} & \textbf{6.1}\tiny{$\pm$1.8} & \textbf{7.3}\tiny{$\pm$3.8} & \textbf{4.0}\tiny{$\pm$1.7} & \textbf{5.7}\tiny{$\pm$2.5}
& \underline{7.3}\tiny{$\pm$1.3} & 7.5\tiny{$\pm$1.2} & \textbf{3.3}\tiny{$\pm$1.4} & \underline{5.1}\tiny{$\pm$0.9} & \underline{6.9}\tiny{$\pm$1.3} \\
\cmidrule{1-12}

\multirow{2}{*}{\textit{Proto.}}
& FeTrIL
& 21.4\tiny{$\pm$7.3} & 40.5\tiny{$\pm$8.1} & 27.8\tiny{$\pm$5.3} & 39.8\tiny{$\pm$9.4} & 39.3\tiny{$\pm$8.1}
& 67.9\tiny{$\pm$10.7} & 67.9\tiny{$\pm$10.5} & 32.9\tiny{$\pm$4.2} & 67.2\tiny{$\pm$10.3} & 66.4\tiny{$\pm$10.9} \\

& FeCAM
& \textbf{5.8}\tiny{$\pm$1.6} & \underline{8.0}\tiny{$\pm$0.9} & \underline{9.8}\tiny{$\pm$2.6} & \underline{5.6}\tiny{$\pm$0.7} & \underline{6.1}\tiny{$\pm$1.2}
& \textbf{3.4}\tiny{$\pm$1.1} & \textbf{3.4}\tiny{$\pm$1.3} & \underline{3.9}\tiny{$\pm$1.4} & \textbf{3.8}\tiny{$\pm$1.2} & \textbf{4.6}\tiny{$\pm$1.5} \\

\midrule
\multicolumn{12}{c}{\textbf{\textit{Plasticity (\%) $\uparrow$}}}\\
\midrule

\textit{LB} & Naive
& 79.1\tiny{$\pm$0.7} & 75.4\tiny{$\pm$0.8} & 54.4\tiny{$\pm$1.6} & 76.3\tiny{$\pm$0.4} & 69.7\tiny{$\pm$0.3}
& 98.4\tiny{$\pm$0.7} & 97.5\tiny{$\pm$0.7} & 53.6\tiny{$\pm$5.2} & 95.8\tiny{$\pm$1.2} & 94.8\tiny{$\pm$0.8} \\
\cmidrule{1-12}

\textit{Reg.} & EWC
& 78.9\tiny{$\pm$0.2} & 74.7\tiny{$\pm$0.9} & \underline{53.1}\tiny{$\pm$1.1} & 76.6\tiny{$\pm$0.1} & 68.6\tiny{$\pm$0.8}
& 98.3\tiny{$\pm$0.6} & 97.4\tiny{$\pm$0.7} & 53.4\tiny{$\pm$5.0} & 95.8\tiny{$\pm$1.2} & 94.8\tiny{$\pm$0.7} \\
\cmidrule{1-12}

\multirow{4}{*}{\textit{Rep.}}
& Replay
& 76.9\tiny{$\pm$0.3} & 73.7\tiny{$\pm$1.1} & 52.0\tiny{$\pm$2.6} & 75.1\tiny{$\pm$0.4} & 68.7\tiny{$\pm$0.6}
& 98.2\tiny{$\pm$0.6} & 97.3\tiny{$\pm$0.7} & \underline{53.6}\tiny{$\pm$4.0} & 95.7\tiny{$\pm$1.1} & 94.6\tiny{$\pm$0.8} \\

& iCaRL
& 76.3\tiny{$\pm$0.6} & 73.6\tiny{$\pm$2.2} & \textbf{53.4}\tiny{$\pm$2.5} & 74.5\tiny{$\pm$0.7} & 68.1\tiny{$\pm$0.8}
& \textbf{98.8}\tiny{$\pm$0.8} & \textbf{98.2}\tiny{$\pm$0.9} & \textbf{54.8}\tiny{$\pm$6.8} & \underline{96.7}\tiny{$\pm$1.4} & \underline{96.0}\tiny{$\pm$1.4} \\

& DER++
& 76.5\tiny{$\pm$1.7} & 73.6\tiny{$\pm$2.6} & 48.8\tiny{$\pm$1.2} & 73.8\tiny{$\pm$1.5} & 67.0\tiny{$\pm$3.1}
& 97.8\tiny{$\pm$0.3} & 96.7\tiny{$\pm$0.3} & 45.2\tiny{$\pm$2.2} & 95.0\tiny{$\pm$0.4} & 94.0\tiny{$\pm$1.4} \\

& RER
& 77.3\tiny{$\pm$0.2} & 74.3\tiny{$\pm$0.7} & 50.4\tiny{$\pm$2.6} & 75.0\tiny{$\pm$0.2} & 68.8\tiny{$\pm$0.9}
& 98.2\tiny{$\pm$0.6} & 97.3\tiny{$\pm$0.7} & 53.1\tiny{$\pm$4.0} & 95.7\tiny{$\pm$1.1} & 94.6\tiny{$\pm$0.8} \\
\cmidrule{1-12}

\textit{Sub.} & EASE
& 64.2\tiny{$\pm$1.4} & 58.4\tiny{$\pm$2.1} & 40.3\tiny{$\pm$2.4} & 62.9\tiny{$\pm$0.6} & 51.8\tiny{$\pm$1.4}
& 66.9\tiny{$\pm$0.8} & 64.2\tiny{$\pm$0.4} & 49.6\tiny{$\pm$0.7} & 64.6\tiny{$\pm$1.2} & 61.4\tiny{$\pm$0.5} \\
\cmidrule{1-12}

\textit{Proj.} & RanPAC
& \textbf{80.1}\tiny{$\pm$1.8} & \underline{75.7}\tiny{$\pm$1.5} & 27.6\tiny{$\pm$3.1} & 76.8\tiny{$\pm$1.4} & 70.5\tiny{$\pm$1.9}
& 88.5\tiny{$\pm$1.1} & 86.3\tiny{$\pm$1.1} & 28.9\tiny{$\pm$1.2} & 89.1\tiny{$\pm$0.8} & 84.8\tiny{$\pm$1.1} \\
\cmidrule{1-12}

\multirow{2}{*}{\textit{Proto.}}
& FeTrIL
& 79.7\tiny{$\pm$0.8} & \textbf{77.2}\tiny{$\pm$0.4} & 52.7\tiny{$\pm$0.3} & \underline{77.0}\tiny{$\pm$0.6} & \underline{73.0}\tiny{$\pm$1.3}
& \underline{98.8}\tiny{$\pm$0.2} & \underline{98.1}\tiny{$\pm$0.4} & 44.3\tiny{$\pm$1.4} & \textbf{97.1}\tiny{$\pm$0.4} & \textbf{97.0}\tiny{$\pm$0.5} \\

& FeCAM
& \underline{80.0}\tiny{$\pm$1.4} & 73.9\tiny{$\pm$0.8} & 51.8\tiny{$\pm$2.2} & \textbf{77.2}\tiny{$\pm$0.5} & \textbf{74.8}\tiny{$\pm$1.0}
& 58.7\tiny{$\pm$0.8} & 54.6\tiny{$\pm$1.1} & 41.0\tiny{$\pm$1.2} & 50.9\tiny{$\pm$1.0} & 34.1\tiny{$\pm$1.3} \\

\bottomrule
\end{tabular}
}
\renewcommand{\arraystretch}{1}
\end{table*}

\begin{figure*}[t]
\centering
\includegraphics[width=\textwidth]{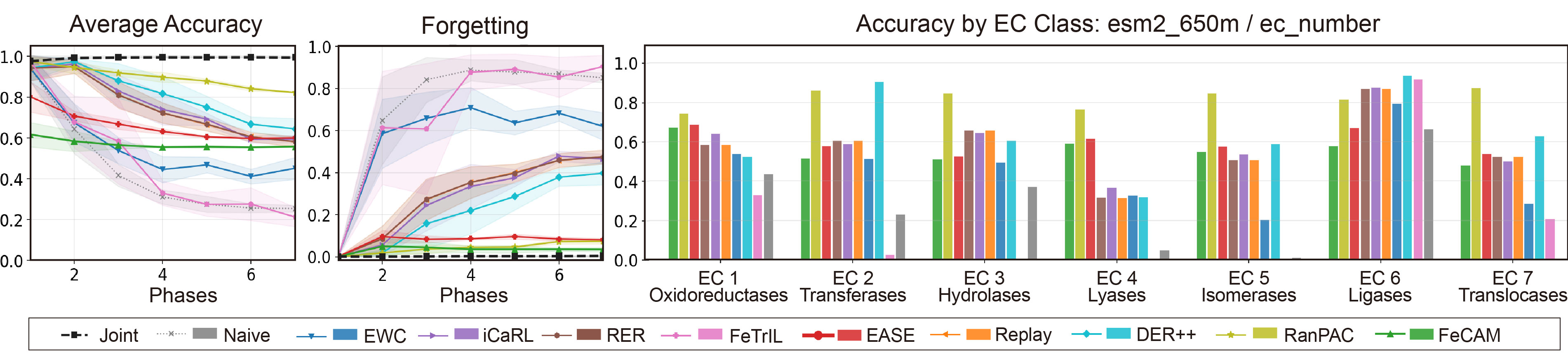}
\caption{Continual-learning performance on EC~Number with ESM2-650M.
\textbf{Left:} Average incremental accuracy across phases.
\textbf{Middle:} Forgetting across phases.
\textbf{Right:} Final accuracy across the seven top-level EC families.}
\label{fig:main_curves}
\end{figure*}

Figure~\ref{fig:main_curves} provides a representative view of the
phase-wise behavior on EC~Number with ESM2-650M. The trajectories show
how the aggregate differences in Table~\ref{tab:main} emerge as classes
are introduced, while the right panel shows that performance also varies
across top-level EC families. Per-encoder trajectories for the remaining
configurations are provided in Appendix
Figs.~\ref{fig:enzyme_reaction_curves} and~\ref{fig:ecnumber_curves}.

\subsection{Robustness Across Phase Construction Protocols}

\paragraph{Robustness to phase construction.}
Table~\ref{tab:baseline_protocol_sanity} examines whether method rankings depend on the EC-first phase construction used in the main experiments. The overall performance pattern remains broadly stable across all three alternative protocols. RanPAC remains the strongest method on EC~Number, while RanPAC and FeCAM are consistently strong on Enzyme Reaction; DER++ also remains competitive across both datasets. EWC and FeTrIL remain among the weakest methods. On Enzyme Reaction, RanPAC and FeCAM achieve 60--63\% accuracy across all protocols. On EC~Number, RanPAC remains near 69\%, whereas FeCAM reaches 44--50\%. Finer-EC gives modestly higher accuracy than Random or Mixed for several methods. For example, FeCAM increases from 44.7\% to 49.5\% on EC~Number. This may reflect reduced inter-phase interference when functionally related classes are grouped together, but it does not alter the broader performance pattern. Forgetting and plasticity also preserve the same qualitative structure across protocols. FeCAM, EASE, and RanPAC maintain low forgetting, whereas most replay- and regularization-based methods show substantially higher forgetting. Large cross-encoder standard deviations also persist across protocols, with RanPAC varying by approximately $\pm$20--28\%. This suggests that the encoder-dependent variability observed in the main experiments is not specific to a particular phase schedule. Overall, the main diagnostic patterns remain robust across different phase constructions. Per-encoder results for the three alternative protocols are reported in Appendix Tables~\ref{tab:baseline_protocol_sanity_per_encoder_a} and~\ref{tab:baseline_protocol_sanity_per_encoder_b}.

\begin{table*}[h]
\centering
\small
\caption{Protocol sanity check with baselines under alternative phase constructions. We report encoder-average Avg Acc, Forgetting, and Plasticity (\%, mean$\pm$std over the five encoders; per-encoder values are averaged over three seeds) under the random, finer-EC, and mixed phase-construction protocols.}
\label{tab:baseline_protocol_sanity}
\vspace{2pt}
\resizebox{\textwidth}{!}{%
\renewcommand{\arraystretch}{1.0}
\begin{tabular}{ll | ccc | ccc | ccc}
\toprule
\multirow{2}{*}{\shortstack[l]{\textbf{Data-}\\\textbf{set}}} 
& \multirow{2}{*}{\textbf{Method}}
& \multicolumn{3}{c|}{\textbf{Random classes}}
& \multicolumn{3}{c|}{\textbf{Finer-EC buckets}}
& \multicolumn{3}{c}{\textbf{Mixed hierarchy}} \\
\cmidrule(lr){3-5} \cmidrule(lr){6-8} \cmidrule(lr){9-11}
&
& \textbf{Acc} $\uparrow$ & \textbf{Fgt} $\downarrow$ & \textbf{Plast} $\uparrow$
& \textbf{Acc} $\uparrow$ & \textbf{Fgt} $\downarrow$ & \textbf{Plast} $\uparrow$
& \textbf{Acc} $\uparrow$ & \textbf{Fgt} $\downarrow$ & \textbf{Plast} $\uparrow$ \\
\midrule

\multirow{9}{*}{\shortstack[l]{\textbf{Enzyme}\\\textbf{Reaction}}}
& EWC
& 14.5\tiny{$\pm$2.8} & 76.3\tiny{$\pm$4.6} & \textbf{79.9}\tiny{$\pm$6.5}
& 14.8\tiny{$\pm$3.1} & 71.5\tiny{$\pm$4.0} & 76.1\tiny{$\pm$6.5}
& 15.4\tiny{$\pm$3.3} & 76.5\tiny{$\pm$3.5} & \textbf{81.0}\tiny{$\pm$6.2} \\

& FeTrIL
& 21.4\tiny{$\pm$2.2} & 68.3\tiny{$\pm$6.4} & \textbf{79.9}\tiny{$\pm$7.6}
& 23.4\tiny{$\pm$3.0} & 62.6\tiny{$\pm$4.8} & \textbf{77.0}\tiny{$\pm$7.2}
& 22.1\tiny{$\pm$2.7} & 68.1\tiny{$\pm$5.5} & 80.5\tiny{$\pm$7.4} \\

& EASE
& 39.2\tiny{$\pm$9.4} & 14.8\tiny{$\pm$0.5} & 51.8\tiny{$\pm$9.4}
& 42.8\tiny{$\pm$9.9} & 12.6\tiny{$\pm$0.7} & 53.5\tiny{$\pm$9.9}
& 39.3\tiny{$\pm$9.3} & 14.5\tiny{$\pm$1.0} & 51.6\tiny{$\pm$9.8} \\

& RER
& 46.6\tiny{$\pm$6.9} & 35.2\tiny{$\pm$2.1} & 76.6\tiny{$\pm$7.1}
& 49.3\tiny{$\pm$8.1} & 29.2\tiny{$\pm$1.5} & 74.1\tiny{$\pm$7.2}
& 49.4\tiny{$\pm$7.3} & 32.8\tiny{$\pm$1.5} & 77.5\tiny{$\pm$7.2} \\

& Replay
& 46.5\tiny{$\pm$6.9} & 35.5\tiny{$\pm$2.0} & 76.8\tiny{$\pm$6.7}
& 48.9\tiny{$\pm$8.2} & 29.8\tiny{$\pm$2.0} & 74.3\tiny{$\pm$6.7}
& 49.1\tiny{$\pm$6.9} & 33.1\tiny{$\pm$0.9} & 77.6\tiny{$\pm$6.7} \\

& iCaRL
& 49.5\tiny{$\pm$7.7} & 30.7\tiny{$\pm$1.8} & 75.7\tiny{$\pm$6.2}
& 53.7\tiny{$\pm$8.7} & 23.9\tiny{$\pm$3.3} & 73.9\tiny{$\pm$6.2}
& 53.5\tiny{$\pm$8.1} & 26.9\tiny{$\pm$2.9} & 76.6\tiny{$\pm$5.9} \\

& DER++
& 55.3\tiny{$\pm$10.5} & 19.9\tiny{$\pm$5.7} & 72.0\tiny{$\pm$6.1}
& 58.3\tiny{$\pm$11.1} & 16.6\tiny{$\pm$5.1} & 71.7\tiny{$\pm$7.1}
& 57.0\tiny{$\pm$9.9} & 19.6\tiny{$\pm$5.1} & 73.5\tiny{$\pm$5.8} \\

& RanPAC
& 60.1\tiny{$\pm$20.1} & 12.6\tiny{$\pm$0.5} & 70.8\tiny{$\pm$19.9}
& \textbf{62.6}\tiny{$\pm$21.9} & \textbf{6.4}\tiny{$\pm$2.0} & 68.0\tiny{$\pm$20.6}
& \textbf{61.5}\tiny{$\pm$20.8} & \textbf{9.8}\tiny{$\pm$0.8} & 69.9\tiny{$\pm$20.4} \\

& FeCAM
& \textbf{60.6}\tiny{$\pm$11.9} & \textbf{12.1}\tiny{$\pm$1.7} & 70.9\tiny{$\pm$12.2}
& 62.0\tiny{$\pm$12.3} & 8.7\tiny{$\pm$1.3} & 69.4\tiny{$\pm$11.7}
& 61.2\tiny{$\pm$12.1} & \textbf{9.8}\tiny{$\pm$0.5} & 70.4\tiny{$\pm$12.6} \\

\midrule

\multirow{9}{*}{\shortstack[l]{\textbf{EC}\\\textbf{Number}}}
& FeTrIL
& 22.7\tiny{$\pm$1.2} & 88.6\tiny{$\pm$2.0} & 98.5\tiny{$\pm$3.0}
& 18.8\tiny{$\pm$1.0} & 93.0\tiny{$\pm$2.3} & 98.5\tiny{$\pm$2.9}
& 25.2\tiny{$\pm$1.4} & 85.3\tiny{$\pm$2.3} & 98.1\tiny{$\pm$4.0} \\

& EWC
& 31.5\tiny{$\pm$8.9} & 78.2\tiny{$\pm$6.9} & 98.6\tiny{$\pm$3.0}
& 37.4\tiny{$\pm$12.6} & 71.3\tiny{$\pm$11.4} & 98.6\tiny{$\pm$2.9}
& 28.4\tiny{$\pm$6.8} & 81.0\tiny{$\pm$2.8} & 97.8\tiny{$\pm$4.6} \\

& FeCAM
& 44.7\tiny{$\pm$10.2} & \textbf{6.5}\tiny{$\pm$0.8} & 50.3\tiny{$\pm$10.8}
& 49.5\tiny{$\pm$10.3} & \textbf{3.7}\tiny{$\pm$0.4} & 52.7\tiny{$\pm$10.4}
& 46.2\tiny{$\pm$10.5} & \textbf{5.5}\tiny{$\pm$0.8} & 50.8\tiny{$\pm$10.9} \\

& RER
& 49.3\tiny{$\pm$8.6} & 58.1\tiny{$\pm$8.0} & 98.7\tiny{$\pm$2.6}
& 53.5\tiny{$\pm$10.7} & 53.1\tiny{$\pm$10.5} & 98.7\tiny{$\pm$2.5}
& 46.5\tiny{$\pm$6.6} & 61.0\tiny{$\pm$5.1} & 98.1\tiny{$\pm$3.9} \\

& Replay
& 49.2\tiny{$\pm$8.4} & 58.4\tiny{$\pm$7.9} & 98.7\tiny{$\pm$2.6}
& 52.6\tiny{$\pm$10.3} & 54.1\tiny{$\pm$10.0} & 98.7\tiny{$\pm$2.5}
& 46.5\tiny{$\pm$6.8} & 61.1\tiny{$\pm$5.3} & 98.1\tiny{$\pm$3.8} \\

& iCaRL
& 55.3\tiny{$\pm$10.3} & 51.4\tiny{$\pm$10.6} & \textbf{99.0}\tiny{$\pm$1.9}
& 56.0\tiny{$\pm$10.9} & 50.6\tiny{$\pm$11.3} & \textbf{99.1}\tiny{$\pm$1.9}
& 53.5\tiny{$\pm$10.7} & 53.4\tiny{$\pm$10.9} & \textbf{98.6}\tiny{$\pm$2.9} \\

& EASE
& 56.8\tiny{$\pm$7.3} & 8.5\tiny{$\pm$1.4} & 64.0\tiny{$\pm$8.3}
& 56.0\tiny{$\pm$5.1} & 5.9\tiny{$\pm$1.2} & 60.7\tiny{$\pm$5.5}
& 58.0\tiny{$\pm$7.0} & 6.6\tiny{$\pm$1.4} & 63.3\tiny{$\pm$8.0} \\

& DER++
& 63.6\tiny{$\pm$12.4} & 41.3\tiny{$\pm$11.9} & 98.5\tiny{$\pm$3.1}
& 65.3\tiny{$\pm$13.4} & 39.0\tiny{$\pm$12.6} & 98.5\tiny{$\pm$3.2}
& 62.2\tiny{$\pm$12.4} & 42.8\tiny{$\pm$11.8} & 98.0\tiny{$\pm$4.2} \\

& RanPAC
& \textbf{68.7}\tiny{$\pm$26.0} & 9.6\tiny{$\pm$2.4} & 76.9\tiny{$\pm$27.5}
& \textbf{69.3}\tiny{$\pm$27.8} & 7.1\tiny{$\pm$1.8} & 75.3\tiny{$\pm$28.9}
& \textbf{68.9}\tiny{$\pm$26.1} & 11.1\tiny{$\pm$2.5} & 78.5\tiny{$\pm$27.6} \\

\bottomrule
\end{tabular}
}
\renewcommand{\arraystretch}{1.0}
\end{table*}

\subsection{Diagnostic Analysis}

\paragraph{Why regularization and geometry assumptions break.}
Table~\ref{tab:diag_AB} isolates two root causes. For the gradient signal (Hypothesis~A), inter-phase conflict cosines are near zero and the Fisher information is highly concentrated in a small fraction of head parameters. With a frozen encoder, there is therefore little conflicting signal for regularization to preserve, consistent with EWC collapsing toward the naive baseline and RER toward plain Replay. For embedding geometry (Hypothesis~B), the within-class covariance condition number spans roughly five orders of magnitude across encoders, while intrinsic dimensionality and class separability also vary substantially with both encoder and dataset. These differences show that the representations induce markedly different class geometries, making a single fixed geometric assumption unlikely to transfer uniformly across encoders.

\begin{table*}[h]
\centering
\caption{Diagnostic statistics for gradient signal (Hypothesis~A) and embedding geometry (Hypothesis~B) across datasets and encoders, averaged over three seeds. Gradient diagnostics report inter-phase conflict cosine and top-1\% Fisher concentration on the MLP head. Geometry diagnostics report embedding dimension, intrinsic rank $r_{95}$, effective rank, within-class covariance condition number, and between/within class separability.}
\label{tab:diag_AB}
\resizebox{\textwidth}{!}{%
\renewcommand{\arraystretch}{1.0}

\begin{tabular}{ll | cc | ccccc}
\toprule
& & \multicolumn{2}{c}{\textbf{Gradient signal (Hypo.A)}} & \multicolumn{5}{c}{\textbf{Embedding geometry (Hypo.B)}} \\
\cmidrule(lr){3-4} \cmidrule(lr){5-9}
\textbf{Dataset} & \textbf{Encoder}
  & \shortstack{Conflict\\cosine}
  & \shortstack{Fisher\\concentration}
  & \shortstack{Embedding\\dimension}
  & \shortstack{Intrinsic\\rank $r_{95}$}
  & \shortstack{Effective\\rank}
  & \shortstack{Condition\\number}
  & \shortstack{Class\\separability} \\
\midrule
\multirow{5}{*}{\shortstack[l]{Enzyme\\Reaction\\(348 classes)}}
  & ESM2-650M    & $-0.047$ & 0.90 & 1280 & 179 & 17.2 & $1.5\!\times\!10^{5}$ & 0.205 \\
  & ESM2-150M    & $-0.044$ & 0.88 & 640  & 60  & 7.5  & $1.6\!\times\!10^{5}$ & 0.203 \\
  & GearNet-Edge & $-0.048$ & 0.96 & 3072 & 84  & 31.7 & $7.1\!\times\!10^{9}$ & 0.241 \\
  & SaProt       & $-0.045$ & 0.78 & 1280 & 177 & 29.4 & $6.9\!\times\!10^{6}$ & 0.209 \\
  & ProtBert     & $-0.046$ & 0.92 & 1024 & 75  & 9.6  & $8.6\!\times\!10^{4}$ & 0.165 \\
\midrule
\multirow{5}{*}{\shortstack[l]{EC Number\\(4788 classes)}}
  & ESM2-650M    & $-0.010$ & 0.94 & 1280 & 217 & 48.1 & $3.8\!\times\!10^{4}$ & 1.167 \\
  & ESM2-150M    & $-0.012$ & 0.96 & 640  & 105 & 37.1 & $1.8\!\times\!10^{4}$ & 1.179 \\
  & GearNet-Edge & $-0.015$ & 0.98 & 3072 & 97  & 36.6 & $3.8\!\times\!10^{9}$ & 0.966 \\
  & SaProt       & $-0.013$ & 0.85 & 1280 & 189 & 49.3 & $3.4\!\times\!10^{6}$ & 0.856 \\
  & ProtBert     & $-0.016$ & 0.96 & 1024 & 114 & 9.8  & $6.8\!\times\!10^{4}$ & 0.450 \\
\bottomrule
\end{tabular}
}
\renewcommand{\arraystretch}{1}
\end{table*}

\begin{figure*}[t]
\centering
\includegraphics[width=\textwidth]{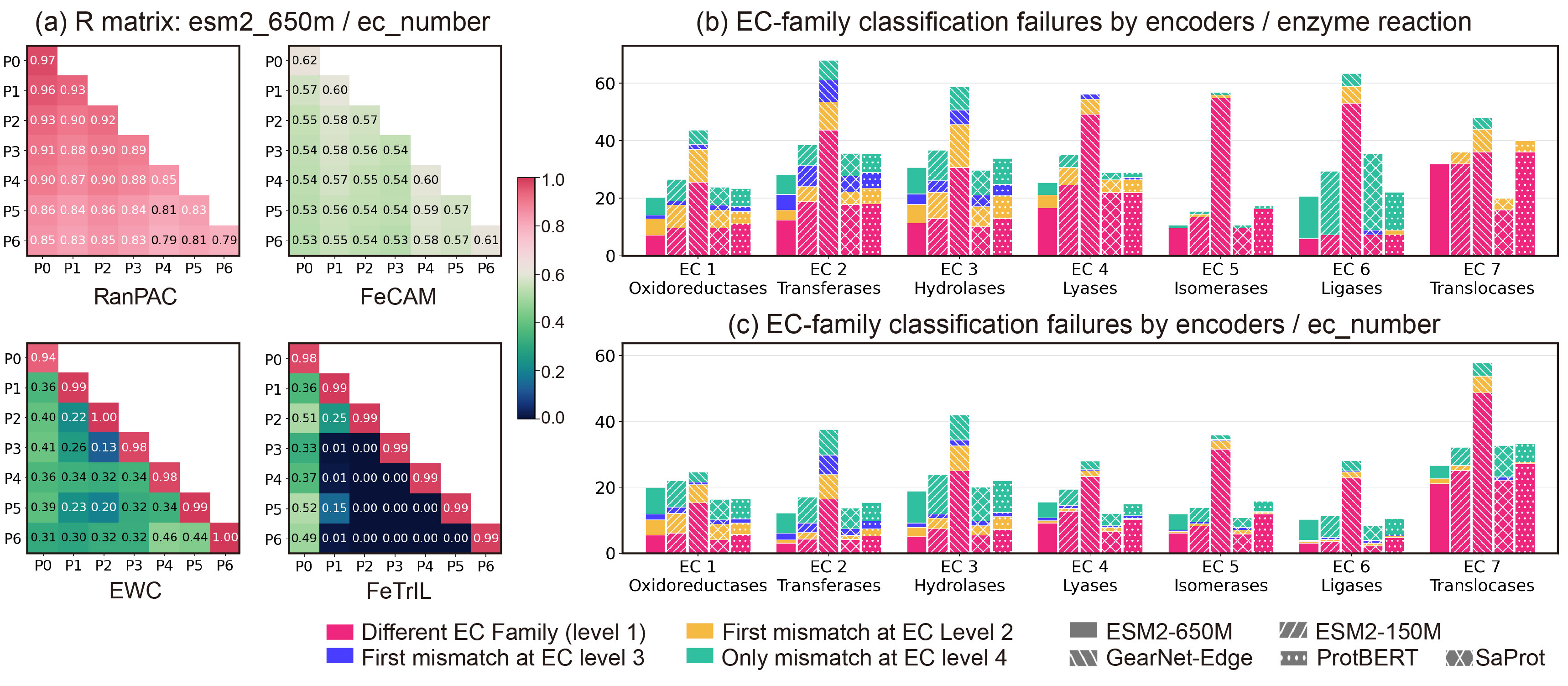}
\caption{Representative diagnostic patterns across methods and protein
encoders. \textbf{(a)} Phase-wise accuracy matrices $R$ for RanPAC,
FeCAM, EWC, and FeTrIL on EC~Number with ESM2-650M.
\textbf{(b)(c)} EC hierarchy error decomposition across encoders on Enzyme
Reaction and EC~Number.}
\label{fig:diagnostic_patterns}
\end{figure*}

The representation dependence is also visible in
Fig.~\ref{fig:diagnostic_patterns}. The phase-wise matrices show distinct
retention patterns among representative methods under the same frozen
representation, while the hierarchy-level error profiles differ across
encoders and datasets. These patterns are consistent with the geometric
differences measured in Table~\ref{tab:diag_AB}. Additional phase-wise
accuracy matrices across encoder--dataset configurations are provided in
Appendix Fig.~\ref{fig:appendix_rmatrix}.

\paragraph{How the shared-covariance assumption fails.}
Table~\ref{tab:confusion} shows that failures under a single shared covariance are structured rather than uniformly distributed across classes. Errors concentrate particularly on transferases and hydrolases, two large and functionally distinct EC families that are repeatedly mapped into one another under the shared metric. Importantly, with the covariance estimator held fixed, the severity of this confusion varies sharply across encoders. Transferase--hydrolase confusions are especially prominent for ESM2-150M, whereas the same failure mode is much weaker for SaProt. This encoder dependence suggests that the limitation is geometric: a single global covariance can be too coarse when different regions of the representation space exhibit substantially different local structure. Moving to finer covariance estimates can alleviate this mismatch for several encoders, but finer granularity is not universally preferable. The full cross-family confusion breakdowns for all five encoders on both datasets are provided in Appendix Table~\ref{tab:full_confusion}.

\begin{table*}[h]
\centering
\caption{Representative cross-family confusions under a single shared covariance on EC~Number.
ESM2-150M and SaProt provide contrasting examples of cross-family error structure across encoders.
Each row reports a true EC family, its predicted family, the number of misclassified instances, and one representative EC-level confusion.}
\label{tab:confusion}
\small
\setlength{\tabcolsep}{4pt}
\renewcommand{\arraystretch}{1.0}
\resizebox{\textwidth}{!}{%
\begin{tabular}{llrl|llrl}
\toprule
\multicolumn{4}{c|}{\textbf{ESM2-150M}}
& \multicolumn{4}{c}{\textbf{SaProt}} \\
True & Pred.\ & \# mis. & Example
& True & Pred.\ & \# mis. & Example \\
\midrule
3 (hydro.)   & 2 (transf.)   & 461 & 3.1.4.35 $\to$ 2.4.1.229
& 2 (transf.) & 3 (hydro.)    & 84 & 2.3.2.27 $\to$ 3.2.2.6 \\
2 (transf.)  & 3 (hydro.)    & 270 & 2.7.2.3 $\to$ 3.5.1.76
& 3 (hydro.)  & 2 (transf.)   & 77 & 3.5.4.26 $\to$ 2.1.1.85 \\
2 (transf.)  & 1 (oxidored.) & 170 & 2.7.1.148 $\to$ 1.1.1.193
& 3 (hydro.)  & 1 (oxidored.) & 39 & 3.5.4.19 $\to$ 1.1.1.23 \\
3 (hydro.)   & 1 (oxidored.) & 159 & 3.1.21.4 $\to$ 1.13.11.58
& 7 (transl.) & 1 (oxidored.) & 33 & 7.1.1.9 $\to$ 1.7.2.5 \\
2 (transf.)  & 4 (lyases)    & 117 & 2.4.2.9 $\to$ 4.2.1.91
& 2 (transf.) & 1 (oxidored.) & 28 & 2.7.2.3 $\to$ 1.2.1.28 \\
2 (transf.)  & 6 (ligases)   & 102 & 2.7.7.101 $\to$ 6.5.1.2
& 2 (transf.) & 4 (lyases)    & 22 & 2.5.1.18 $\to$ 4.4.1.20 \\
\bottomrule
\end{tabular}%
}
\end{table*}

Table~\ref{tab:diag_C} probes this trade-off by comparing three covariance-free controls with four Mahalanobis variants of increasing covariance granularity. Euclidean and Cosine NCM perform substantially below the Mahalanobis variants on both datasets, while Ridge also degrades sharply on EC~Number. Among the Mahalanobis variants, the preferred granularity differs across datasets and encoders. On Enzyme Reaction, full shared covariance is best for all five encoders, outperforming per-class covariance by 6.3--22.4\%. On EC~Number, per-class covariance is best for four of five encoders, improving over full shared covariance by 3.4--6.5\%. GearNet is the consistent exception: full shared covariance improves accuracy from 37.1\% to 47.2\% on Enzyme Reaction and from 58.5\% to 72.4\% on EC~Number. Together, these results show that the appropriate covariance capacity depends jointly on dataset scale and encoder-induced geometry.

\begin{table*}[h]
\centering
\caption{Effect of covariance granularity on prototype classification across datasets and encoders.
Classification accuracy is reported on 256-dimensional PCA projections for three covariance-free controls and four Mahalanobis variants, averaged over three seeds.
Best result for each dataset--encoder pair is shown in \textbf{bold}.}
\label{tab:diag_C}
\small
\resizebox{\textwidth}{!}{%
\renewcommand{\arraystretch}{1.05}
\begin{tabular}{ll | ccc | cccc}
\toprule
& & \multicolumn{3}{c|}{\textbf{No covariance assumption}} & \multicolumn{4}{c}{\textbf{Mahalanobis (covariance granularity)}} \\
\cmidrule(lr){3-5} \cmidrule(lr){6-9}
\textbf{Dataset} & \textbf{Encoder}
  & \shortstack{Euclidean\\NCM}
  & \shortstack{Cosine\\NCM}
  & \shortstack{Ridge\\classifier}
  & \shortstack{Shared\\(diagonal)}
  & \shortstack{Full\\shared}
  & \shortstack{Grouped\\$K\!=\!8$}
  & \shortstack{Per-\\class} \\
\midrule
\multirow{5}{*}{\shortstack[l]{Enzyme\\Reaction\\(348 classes)}}
  & ESM2-650M    & 0.130 & 0.167 & 0.506 & 0.594 & \textbf{0.610} & 0.490 & 0.541 \\
  & ESM2-150M    & 0.108 & 0.147 & 0.484 & 0.586 & \textbf{0.591} & 0.478 & 0.523 \\
  & GearNet-Edge & 0.183 & 0.198 & 0.295 & 0.371 & \textbf{0.472} & 0.267 & 0.248 \\
  & SaProt       & 0.220 & 0.236 & 0.474 & 0.603 & \textbf{0.615} & 0.566 & 0.552 \\
  & ProtBert     & 0.058 & 0.074 & 0.451 & 0.576 & \textbf{0.604} & 0.451 & 0.473 \\
\midrule
\multirow{5}{*}{\shortstack[l]{EC Number\\(4788 classes)}}
  & ESM2-650M    & 0.436 & 0.426 & 0.340 & 0.760 & 0.785 & 0.734 & \textbf{0.849} \\
  & ESM2-150M    & 0.468 & 0.451 & 0.334 & 0.741 & 0.768 & 0.702 & \textbf{0.833} \\
  & GearNet-Edge & 0.372 & 0.381 & 0.207 & 0.585 & \textbf{0.724} & 0.553 & 0.563 \\
  & SaProt       & 0.429 & 0.412 & 0.327 & 0.752 & 0.796 & 0.730 & \textbf{0.830} \\
  & ProtBert     & 0.206 & 0.225 & 0.297 & 0.712 & 0.762 & 0.699 & \textbf{0.806} \\
\bottomrule
\end{tabular}
}
\renewcommand{\arraystretch}{1.0}
\end{table*}

\section{Discussion and Conclusion}

Our results show that class-incremental learning on frozen protein representations is not primarily a matter of choosing a stronger existing CIL baseline. Across five encoders and two dataset scales, no method is consistently reliable, and the same strategy can move from near-best to poor performance when only the encoder changes. The diagnostic analysis identifies three recurring causes. First, gradient-based regularization receives little useful stability signal once the encoder is frozen: inter-phase gradient conflict remains near zero and Fisher information is highly concentrated in a small fraction of classifier parameters. Second, protein encoders induce markedly different embedding geometries, with large variation in intrinsic dimensionality, covariance conditioning, and class separability. Third, the statistical capacity required by prototype classifiers changes with both dataset scale and encoder geometry. These findings suggest that the central difficulty is not catastrophic forgetting alone, but a mismatch between the assumptions built into existing CIL methods and the representation on which they operate.

This distinguishes frozen protein CIL from the vision settings in which most existing CIL methods, including recent approaches built on pretrained representations, were developed and evaluated~\cite{masana2022class,zhou2024continual,zhou2023revisiting}. Our benchmark spans sequence language models, a structure-based graph network, and a sequence--structure fusion model~\cite{lin2023evolutionary,zhang2023protein,su2024saprot}, which encode different sources of protein information and produce substantially different representation geometries. A geometric assumption that is appropriate for one encoder therefore need not transfer to another. RanPAC, for example, is highly competitive on several encoders but deteriorates sharply on GearNet, while FeCAM preserves old classes extremely well yet loses plasticity as its shared-covariance assumption becomes too restrictive. The phase-construction experiments reinforce this conclusion: changing how classes are distributed across phases alters absolute performance only modestly, while the broader encoder-dependent patterns remain. The failures are therefore representation-dependent rather than driven by phase construction.

The covariance analysis provides a concrete example of this broader problem. A single shared covariance offers stable estimation but can be too coarse to represent heterogeneous class geometry, whereas per-class covariance is more flexible but relies on substantially fewer samples per estimate. This bias--variance tension is particularly relevant in high-dimensional covariance estimation, where shrinkage is commonly used to stabilize estimates under limited samples~\cite{ledoit2004well,chen2010shrinkage}. Neither extreme is uniformly preferable in our experiments. Full shared covariance is best across all five encoders on Enzyme Reaction, while per-class covariance is best for four of five encoders on EC~Number; GearNet remains the exception, preferring shared covariance even at the larger scale. This suggests a natural direction for protein CIL: covariance capacity should be adapted between these two extremes rather than fixed in advance. Classes with compatible geometry could share statistical strength, while geometrically distinct regions could retain separate covariance structure. Such adaptation could also incorporate the EC hierarchy as a biological prior, while allowing representation geometry to determine when that hierarchy should or should not define the statistical grouping.

Our findings point toward encoder-aware continual learning as a more suitable design principle for protein representations. Covariance granularity is one instance of this idea, but the same principle may extend to projection dimension, distance metrics, regularization strength, and memory allocation. Rather than searching for a vision-derived CIL recipe that transfers unchanged across protein encoders, future methods should adapt their statistical capacity to the geometry of the representation they receive. In this sense, the benchmark provides not only a negative transfer result but also a design criterion: reliable protein CIL should respond to encoder geometry instead of assuming it away.

Our study deliberately isolates continual-learning behavior under frozen protein representations, and the conclusions should therefore be interpreted within this setting. Partially or fully adapted encoders may exhibit different stability and geometric properties and remain an important direction for future work. Our evaluation is also limited to enzyme function classification under the EC hierarchy, and whether the same failure modes extend to other protein tasks remains open. In addition, our fixed-hyperparameter protocol is designed to evaluate plug-and-play transfer across encoders rather than the best performance achievable through encoder-specific tuning. These choices define the scope of the study rather than weaknesses of the benchmark, and they help separate representation-dependent failure modes from gains that could arise through additional adaptation or tuning.

\clearpage
{\small
\bibliographystyle{unsrtnat}
\bibliography{refs}
}

\clearpage
\appendix
\section*{Declaration of LLM Usage}
Large language models were used solely as writing and coding assistants
during the preparation of this work. Specifically, they were used to
polish prose, improve clarity, and assist with routine coding tasks such
as debugging, result formatting, and figure generation. All experimental
design, analyses, results, and conclusions were determined and verified
by the authors.

\section*{Ethics Statement}
This work evaluates class-incremental learning methods on publicly
available protein sequence and function datasets. It involves no human
subjects, personally identifiable information, or newly collected
biological samples. The study is methodological and diagnostic in nature,
with the aim of understanding how continual-learning methods interact
with different frozen protein representations. We are not aware of
foreseeable harmful applications arising directly from this work.

\section*{Reproducibility Statement}
We provide anonymized code at
\url{https://anonymous.4open.science/r/ProteinCIL/}.
Dataset construction, preprocessing, phase protocols, and encoder details
are documented in Appendix~\ref{app:data_pipeline}, while experimental
settings and baseline configurations are provided in
Appendix~\ref{app:setup}. All protein encoders are kept frozen throughout
continual learning, and experiments are evaluated under consistent
protocols across representations. Results are averaged over three random
seeds unless otherwise specified. All datasets and pretrained encoders
used in this work are publicly available and cited to their original
sources. All experiments were run on a single NVIDIA RTX 4090 GPU.

\appendix


\section{Benchmark Details}
\label{app:data_pipeline}

\subsection{Datasets and Preprocessing}
\label{app:datasets}

Table~\ref{tab:app_datasets} summarizes the two datasets and the preprocessing steps used in our benchmark. Both datasets are converted to a unified format. Each sample contains an amino acid sequence, a functional label, and an associated protein structure when available. For Enzyme Reaction, multi-label EC annotations are converted to a single label using argmax, and the original training and validation splits are merged. For EC Number, only the first annotation is retained for proteins with multiple labels. AlphaFold v4 predicted structures are downloaded for EC Number. Enzyme Reaction structures are obtained from the TorchDrug distribution. Samples without matched structures are excluded from GearNet-Edge and SaProt but retained for sequence-only encoders. All PDB files are processed with Foldseek~\citep{van2024fast} to generate the per-residue 3Di structural tokens required by SaProt.

\begin{table*}[h]
\centering
\caption{Dataset statistics. Counts report the final cached sequence-encoder splits; structure-dependent encoders may exclude samples without matched structures.}
\label{tab:app_datasets}
\small
\begin{tabular*}{\textwidth}{@{\extracolsep{\fill}}lccccc@{}}
\toprule
Dataset & Classes & Train & Test & Train/class & Phases \\
\midrule
Enzyme Reaction & 348 & 29{,}215 & 5{,}651 & $\sim$84 & 7 \\
EC Number & 4{,}788 & 73{,}259 & 9{,}112 & $\sim$15 & 7 \\
\bottomrule
\end{tabular*}
\end{table*}

\subsection{Encoders and Embedding Extraction}
\label{app:encoder_extraction}

Table~\ref{tab:app_encoders} lists the five frozen encoders and their extraction settings. Each encoder is applied once to every protein, and the resulting embeddings are cached and reused across all CIL methods. ESM2 models use mean pooling over non-special token representations. ProtBert and SaProt use attention-weighted mean pooling. SaProt~\citep{su2024saprot} combines each amino acid with its corresponding Foldseek 3Di structural token. Proteins without matched 3Di sequences are excluded from SaProt. GearNet-Edge~\citep{zhang2023protein} constructs multi-relational residue graphs from 3D coordinates. Its embeddings are L2-normalized to reduce the effect of magnitude differences on distance-based computations. Sequences that exceed encoder-specific length limits are truncated.

\begin{table*}[h]
\centering
\caption{Frozen encoders and embedding extraction details.}
\label{tab:app_encoders}
\small
\begin{tabular*}{\textwidth}{@{\extracolsep{\fill}}lllcl@{}}
\toprule
Encoder & Type & Input & Dim. & Pooling \\
\midrule
ESM2-650M~\citep{lin2023evolutionary} & Sequence PLM & AA sequence & 1280 & Mean pool, final layer \\
ESM2-150M~\citep{lin2023evolutionary} & Sequence PLM & AA sequence & 640 & Mean pool, final layer \\
ProtBert~\citep{elnaggar2022prottrans} & Sequence PLM & Space-separated AA & 1024 & Attention-weighted mean \\
SaProt~\citep{su2024saprot} & Seq.+Struct. & AA + 3Di tokens & 1280 & Attention-weighted mean \\
GearNet-Edge~\citep{zhang2023protein} & Structure GNN & Residue graph from 3D coords & 3072 & Node mean; L2 norm. \\
\bottomrule
\end{tabular*}
\end{table*}

\subsection{Baselines and Hyperparameters}
\label{app:hyperparams}

Table~\ref{tab:app_baselines} lists the baseline methods and their configurations. All MLP-based methods use a two-layer classifier with hidden dimension 1024. They are trained with Adam at a learning rate of $10^{-3}$ for 100 epochs per phase. Replay, iCaRL, DER++, and RER use a reservoir buffer of 2{,}000 samples. DER++ uses $\alpha{=}0.5$ and $\beta{=}0.5$. EWC uses a Fisher penalty weight of $\lambda{=}2000$. FeCAM uses a global shared covariance with ridge regularization ($\lambda{=}10^{-4}$). FeTrIL generates pseudo-features for old classes by translating new-class embeddings using the difference between old and new class means. It then retrains an MLP classifier on the combined real and pseudo features. RanPAC uses a fixed Gaussian random projection with $D{=}12800$ and ridge parameter $\lambda{=}1.0$. EASE trains a rank-256 low-rank adapter for each phase using the same learning rate and epoch count as the MLP-based methods. All hyperparameters are fixed across encoder-dataset configurations, with no setting-specific tuning.

\begin{table*}[h]
\centering
\caption{Baseline CIL methods and configurations.}
\label{tab:app_baselines}
\small
\resizebox{\textwidth}{!}{%
\begin{tabular}{@{}llll@{}}
\toprule
Method & Strategy & Core Mechanism & Key Hyperparameters \\
\midrule
Naive & Lower bound & Train on new phase only & Hidden=1024, lr=$10^{-3}$, 100 epochs \\
Joint MLP & Joint reference & Retrain on all data & Same as Naive \\
EWC~\citep{kirkpatrick2017overcoming} & Regularization & Fisher-weighted L2 penalty & $\lambda{=}2000$ \\
Replay~\citep{chaudhry2019tiny} & Replay & Fixed-size reservoir buffer & Buffer=2{,}000 \\
iCaRL~\citep{rebuffi2017icarl} & Replay + Distillation & Exemplar replay + distillation & Buffer=2{,}000 \\
DER++~\citep{buzzega2020dark} & Replay + Distillation & Logit replay + label replay & Buffer=2{,}000, $\alpha{=}0.5$, $\beta{=}0.5$ \\
RER~\citep{lin2024rer} & Adaptive replay & Gradient-cosine adaptive $\alpha$ & Buffer=2{,}000 \\
FeTrIL~\citep{petit2023fetril} & Pseudo-feature & Class-mean translation + MLP classifier & Hidden=1024, lr=$10^{-3}$, 100 epochs \\
RanPAC~\citep{mcdonnell2023ranpac} & Projection & Random projection + ridge regression & $D{=}12800$, $\lambda{=}1.0$ \\
EASE~\citep{wang2024ease} & Subspace & Learned per-task adapter + prototypes & Rank=256, lr=$10^{-3}$, 100 epochs \\
FeCAM~\citep{goswami2024fecam} & Prototype & Shared Mahalanobis distance & Ridge $\lambda{=}10^{-4}$ \\
\bottomrule
\end{tabular}
}
\end{table*}

\clearpage
\section{Experimental Details}
\label{app:setup}

\subsection{Per-Encoder Breakdown}
\label{sec:per_encoder_breakdown}

Tables~\ref{tab:baseline_protocol_sanity_per_encoder_a}
and~\ref{tab:baseline_protocol_sanity_per_encoder_b} report the per-encoder
results underlying the encoder-averaged values in
Table~\ref{tab:baseline_protocol_sanity}. Each block covers one encoder on
both datasets, with values averaged over three seeds. Across encoders,
changing the phase construction produces only modest within-encoder
variation, while substantial differences between encoders persist.
On EC Number, replay- and regularization-based methods generally retain
high plasticity but incur substantial forgetting, whereas FeCAM and EASE
maintain much lower forgetting with reduced plasticity. These results
support the main finding that phase construction has a smaller effect than
the choice of frozen protein representation.

\begin{table*}[h]
\centering
\scriptsize
\setlength{\tabcolsep}{3pt}
\renewcommand{\arraystretch}{1.0}
\caption{Per-encoder breakdown of the protocol sanity check (part 1/2: ESM2-150M, ESM2-650M), in the same format as Table~\ref{tab:baseline_protocol_sanity} but reported separately for each encoder instead of averaged over them. Each encoder block contains both datasets. Values are mean$\pm$std over three seeds (\%). Acc = Avg Acc ($\uparrow$), Fgt = Forgetting ($\downarrow$), Plast = Plasticity ($\uparrow$).}
\label{tab:baseline_protocol_sanity_per_encoder_a}
\vspace{2pt}
\resizebox{\textwidth}{!}{%
\begin{tabular}{ll | ccc | ccc | ccc}
\toprule
\multirow{2}{*}{\textbf{Dataset}}
& \multirow{2}{*}{\textbf{Method}}
& \multicolumn{3}{c|}{\textbf{Random classes}}
& \multicolumn{3}{c|}{\textbf{Finer-EC buckets}}
& \multicolumn{3}{c}{\textbf{Mixed hierarchy}} \\
\cmidrule(lr){3-5} \cmidrule(lr){6-8} \cmidrule(lr){9-11}
& 
& \textbf{Acc} $\uparrow$ & \textbf{Fgt} $\downarrow$ & \textbf{Plast} $\uparrow$
& \textbf{Acc} $\uparrow$ & \textbf{Fgt} $\downarrow$ & \textbf{Plast} $\uparrow$
& \textbf{Acc} $\uparrow$ & \textbf{Fgt} $\downarrow$ & \textbf{Plast} $\uparrow$ \\
\midrule
\multicolumn{11}{c}{\textbf{\textit{ESM2-150M}}} \\
\midrule
\multirow{9}{*}{\shortstack[l]{\textbf{Enzyme}\\\textbf{Reaction}}}
& EWC & 14.8\tiny{$\pm$0.8} & 78.9\tiny{$\pm$1.8} & 82.4\tiny{$\pm$2.3} & 15.7\tiny{$\pm$2.5} & 73.2\tiny{$\pm$3.0} & 78.4\tiny{$\pm$0.1} & 15.8\tiny{$\pm$1.8} & 78.0\tiny{$\pm$2.0} & 82.6\tiny{$\pm$0.5} \\
& FeTrIL & 22.4\tiny{$\pm$5.0} & 70.7\tiny{$\pm$6.8} & 83.0\tiny{$\pm$1.7} & 24.4\tiny{$\pm$5.2} & 62.9\tiny{$\pm$6.0} & 78.3\tiny{$\pm$0.4} & 23.3\tiny{$\pm$4.9} & 69.9\tiny{$\pm$5.0} & 83.2\tiny{$\pm$0.6} \\
& EASE & 41.8\tiny{$\pm$1.9} & 14.2\tiny{$\pm$1.6} & 53.8\tiny{$\pm$1.7} & 45.1\tiny{$\pm$1.2} & 11.4\tiny{$\pm$1.8} & 54.7\tiny{$\pm$0.7} & 41.0\tiny{$\pm$4.1} & 14.6\tiny{$\pm$3.9} & 53.5\tiny{$\pm$1.5} \\
& RER & 47.0\tiny{$\pm$3.4} & 36.8\tiny{$\pm$2.6} & 78.5\tiny{$\pm$2.2} & 52.8\tiny{$\pm$1.6} & 28.0\tiny{$\pm$2.1} & 76.7\tiny{$\pm$0.4} & 51.6\tiny{$\pm$2.6} & 33.1\tiny{$\pm$2.0} & 80.0\tiny{$\pm$1.0} \\
& Replay & 46.5\tiny{$\pm$3.4} & 37.1\tiny{$\pm$2.8} & 78.2\tiny{$\pm$2.6} & 52.5\tiny{$\pm$1.8} & 28.5\tiny{$\pm$2.1} & 76.9\tiny{$\pm$0.2} & 50.7\tiny{$\pm$2.0} & 33.3\tiny{$\pm$1.4} & 79.3\tiny{$\pm$1.2} \\
& iCaRL & 52.0\tiny{$\pm$3.2} & 30.1\tiny{$\pm$3.1} & 77.2\tiny{$\pm$3.5} & 56.7\tiny{$\pm$1.0} & 22.7\tiny{$\pm$1.7} & 75.9\tiny{$\pm$0.6} & 54.5\tiny{$\pm$4.5} & 27.2\tiny{$\pm$3.8} & 77.9\tiny{$\pm$1.3} \\
& DER++ & 57.6\tiny{$\pm$3.1} & 19.4\tiny{$\pm$0.5} & 74.2\tiny{$\pm$3.1} & 61.6\tiny{$\pm$5.3} & 15.3\tiny{$\pm$2.8} & 73.9\tiny{$\pm$2.3} & 58.4\tiny{$\pm$2.8} & 19.9\tiny{$\pm$1.9} & 75.2\tiny{$\pm$1.8} \\
& RanPAC & 68.3\tiny{$\pm$3.1} & 11.9\tiny{$\pm$2.6} & 78.4\tiny{$\pm$2.0} & 70.8\tiny{$\pm$0.3} & 6.5\tiny{$\pm$2.0} & 76.4\tiny{$\pm$1.4} & 70.9\tiny{$\pm$0.9} & 9.0\tiny{$\pm$0.4} & 78.6\tiny{$\pm$0.8} \\
& FeCAM & 62.3\tiny{$\pm$4.5} & 14.6\tiny{$\pm$7.7} & 74.8\tiny{$\pm$2.5} & 63.4\tiny{$\pm$1.1} & 9.6\tiny{$\pm$2.2} & 71.7\tiny{$\pm$1.0} & 64.2\tiny{$\pm$3.3} & 10.8\tiny{$\pm$3.7} & 73.4\tiny{$\pm$0.1} \\
\cmidrule(l){1-11}
\multirow{9}{*}{\shortstack[l]{\textbf{EC}\\\textbf{Number}}}
& EWC & 35.7\tiny{$\pm$3.1} & 75.0\tiny{$\pm$3.6} & 99.9\tiny{$\pm$0.0} & 43.5\tiny{$\pm$6.4} & 65.8\tiny{$\pm$7.4} & 99.9\tiny{$\pm$0.0} & 30.6\tiny{$\pm$5.2} & 80.9\tiny{$\pm$6.1} & 99.9\tiny{$\pm$0.0} \\
& FeTrIL & 23.1\tiny{$\pm$7.6} & 89.5\tiny{$\pm$8.9} & 99.8\tiny{$\pm$0.1} & 19.2\tiny{$\pm$4.7} & 94.1\tiny{$\pm$5.4} & 99.9\tiny{$\pm$0.0} & 26.0\tiny{$\pm$10.0} & 86.1\tiny{$\pm$12.0} & 99.8\tiny{$\pm$0.2} \\
& EASE & 60.9\tiny{$\pm$0.6} & 8.4\tiny{$\pm$1.2} & 68.0\tiny{$\pm$1.6} & 60.1\tiny{$\pm$0.2} & 5.6\tiny{$\pm$3.8} & 64.6\tiny{$\pm$2.9} & 62.0\tiny{$\pm$0.8} & 6.2\tiny{$\pm$1.9} & 67.3\tiny{$\pm$1.4} \\
& RER & 52.5\tiny{$\pm$2.9} & 55.2\tiny{$\pm$3.4} & 99.9\tiny{$\pm$0.1} & 58.2\tiny{$\pm$4.9} & 48.6\tiny{$\pm$5.7} & 99.8\tiny{$\pm$0.1} & 49.6\tiny{$\pm$2.9} & 58.7\tiny{$\pm$3.4} & 99.9\tiny{$\pm$0.0} \\
& Replay & 52.1\tiny{$\pm$4.9} & 55.8\tiny{$\pm$5.7} & 99.9\tiny{$\pm$0.0} & 57.4\tiny{$\pm$5.1} & 49.5\tiny{$\pm$6.0} & 99.8\tiny{$\pm$0.0} & 48.9\tiny{$\pm$2.8} & 59.4\tiny{$\pm$3.2} & 99.8\tiny{$\pm$0.1} \\
& iCaRL & 60.1\tiny{$\pm$4.2} & 46.5\tiny{$\pm$5.0} & 99.9\tiny{$\pm$0.0} & 61.3\tiny{$\pm$2.1} & 45.0\tiny{$\pm$2.4} & 99.9\tiny{$\pm$0.0} & 58.4\tiny{$\pm$4.9} & 48.4\tiny{$\pm$5.7} & 99.9\tiny{$\pm$0.0} \\
& DER++ & 68.4\tiny{$\pm$4.4} & 36.8\tiny{$\pm$5.2} & 99.9\tiny{$\pm$0.0} & 71.0\tiny{$\pm$3.8} & 33.8\tiny{$\pm$4.4} & 99.9\tiny{$\pm$0.0} & 66.8\tiny{$\pm$3.3} & 38.6\tiny{$\pm$3.9} & 99.9\tiny{$\pm$0.0} \\
& RanPAC & 78.6\tiny{$\pm$0.5} & 12.0\tiny{$\pm$2.1} & 88.9\tiny{$\pm$2.0} & 80.5\tiny{$\pm$0.1} & 8.7\tiny{$\pm$1.9} & 87.9\tiny{$\pm$1.6} & 79.6\tiny{$\pm$1.3} & 12.8\tiny{$\pm$2.9} & 90.6\tiny{$\pm$2.5} \\
& FeCAM & 51.7\tiny{$\pm$1.3} & 7.2\tiny{$\pm$1.9} & 57.9\tiny{$\pm$2.6} & 56.2\tiny{$\pm$0.1} & 3.9\tiny{$\pm$1.0} & 59.5\tiny{$\pm$1.0} & 53.8\tiny{$\pm$1.2} & 5.9\tiny{$\pm$1.5} & 58.8\tiny{$\pm$0.7} \\
\midrule
\multicolumn{11}{c}{\textbf{\textit{ESM2-650M}}} \\
\midrule
\multirow{9}{*}{\shortstack[l]{\textbf{Enzyme}\\\textbf{Reaction}}}
& EWC & 16.2\tiny{$\pm$1.1} & 79.6\tiny{$\pm$0.7} & 84.4\tiny{$\pm$1.7} & 17.3\tiny{$\pm$1.2} & 75.2\tiny{$\pm$2.5} & 81.8\tiny{$\pm$0.9} & 18.4\tiny{$\pm$2.8} & 79.7\tiny{$\pm$3.0} & 86.7\tiny{$\pm$0.3} \\
& FeTrIL & 22.9\tiny{$\pm$4.9} & 73.6\tiny{$\pm$5.9} & 85.6\tiny{$\pm$0.7} & 25.7\tiny{$\pm$5.7} & 66.5\tiny{$\pm$6.9} & 82.7\tiny{$\pm$0.3} & 23.8\tiny{$\pm$5.3} & 73.1\tiny{$\pm$5.8} & 86.5\tiny{$\pm$0.5} \\
& EASE & 47.2\tiny{$\pm$2.5} & 15.1\tiny{$\pm$3.4} & 60.1\tiny{$\pm$0.5} & 49.9\tiny{$\pm$1.5} & 12.8\tiny{$\pm$2.2} & 60.8\tiny{$\pm$0.4} & 47.2\tiny{$\pm$4.2} & 15.9\tiny{$\pm$3.6} & 60.8\tiny{$\pm$1.2} \\
& RER & 53.8\tiny{$\pm$3.8} & 33.1\tiny{$\pm$2.2} & 82.1\tiny{$\pm$1.9} & 56.0\tiny{$\pm$1.2} & 27.8\tiny{$\pm$0.7} & 79.8\tiny{$\pm$0.6} & 56.4\tiny{$\pm$2.0} & 31.1\tiny{$\pm$1.3} & 83.1\tiny{$\pm$1.3} \\
& Replay & 54.0\tiny{$\pm$3.9} & 32.8\tiny{$\pm$2.7} & 82.1\tiny{$\pm$1.7} & 55.4\tiny{$\pm$0.9} & 28.2\tiny{$\pm$1.1} & 79.6\tiny{$\pm$0.4} & 55.8\tiny{$\pm$1.8} & 31.7\tiny{$\pm$0.7} & 83.0\tiny{$\pm$1.3} \\
& iCaRL & 56.1\tiny{$\pm$3.7} & 29.4\tiny{$\pm$3.9} & 81.3\tiny{$\pm$0.4} & 60.7\tiny{$\pm$1.1} & 21.4\tiny{$\pm$1.6} & 79.0\tiny{$\pm$0.4} & 59.7\tiny{$\pm$2.3} & 25.2\tiny{$\pm$0.8} & 81.3\tiny{$\pm$1.8} \\
& DER++ & 65.3\tiny{$\pm$3.4} & 14.8\tiny{$\pm$1.1} & 77.9\tiny{$\pm$2.7} & 68.6\tiny{$\pm$3.6} & 12.0\tiny{$\pm$1.1} & 78.6\tiny{$\pm$2.7} & 65.9\tiny{$\pm$4.4} & 16.2\tiny{$\pm$3.1} & 79.4\tiny{$\pm$1.7} \\
& RanPAC & 72.9\tiny{$\pm$3.0} & 12.7\tiny{$\pm$4.3} & 83.8\tiny{$\pm$1.0} & 75.6\tiny{$\pm$0.1} & 6.3\tiny{$\pm$1.7} & 80.8\tiny{$\pm$1.5} & 74.7\tiny{$\pm$1.1} & 9.5\tiny{$\pm$0.6} & 82.9\tiny{$\pm$0.7} \\
& FeCAM & 69.1\tiny{$\pm$4.2} & 11.9\tiny{$\pm$5.6} & 79.0\tiny{$\pm$0.9} & 70.0\tiny{$\pm$0.5} & 9.2\tiny{$\pm$1.5} & 77.9\tiny{$\pm$0.8} & 69.4\tiny{$\pm$2.6} & 11.3\tiny{$\pm$3.8} & 79.1\tiny{$\pm$0.7} \\
\cmidrule(l){1-11}
\multirow{9}{*}{\shortstack[l]{\textbf{EC}\\\textbf{Number}}}
& EWC & 34.9\tiny{$\pm$2.0} & 75.8\tiny{$\pm$2.3} & 99.9\tiny{$\pm$0.0} & 46.7\tiny{$\pm$4.1} & 62.1\tiny{$\pm$4.8} & 99.9\tiny{$\pm$0.0} & 30.7\tiny{$\pm$5.1} & 80.7\tiny{$\pm$6.0} & 99.8\tiny{$\pm$0.1} \\
& FeTrIL & 23.4\tiny{$\pm$7.8} & 89.3\tiny{$\pm$9.1} & 99.9\tiny{$\pm$0.1} & 19.2\tiny{$\pm$4.7} & 94.0\tiny{$\pm$5.4} & 99.8\tiny{$\pm$0.1} & 26.1\tiny{$\pm$10.2} & 86.1\tiny{$\pm$11.9} & 99.9\tiny{$\pm$0.0} \\
& EASE & 63.1\tiny{$\pm$0.7} & 9.6\tiny{$\pm$0.7} & 71.3\tiny{$\pm$0.6} & 59.5\tiny{$\pm$0.2} & 7.1\tiny{$\pm$3.9} & 65.0\tiny{$\pm$3.1} & 63.9\tiny{$\pm$0.6} & 7.4\tiny{$\pm$2.8} & 70.1\tiny{$\pm$3.2} \\
& RER & 56.6\tiny{$\pm$1.3} & 50.6\tiny{$\pm$1.5} & 99.9\tiny{$\pm$0.0} & 62.0\tiny{$\pm$6.0} & 44.2\tiny{$\pm$7.0} & 99.9\tiny{$\pm$0.0} & 50.8\tiny{$\pm$3.6} & 57.2\tiny{$\pm$4.2} & 99.8\tiny{$\pm$0.1} \\
& Replay & 56.1\tiny{$\pm$1.3} & 51.2\tiny{$\pm$1.5} & 99.9\tiny{$\pm$0.0} & 59.8\tiny{$\pm$5.2} & 46.7\tiny{$\pm$6.1} & 99.9\tiny{$\pm$0.0} & 50.6\tiny{$\pm$3.4} & 57.4\tiny{$\pm$4.0} & 99.8\tiny{$\pm$0.1} \\
& iCaRL & 63.9\tiny{$\pm$3.8} & 42.0\tiny{$\pm$4.4} & 99.9\tiny{$\pm$0.0} & 62.6\tiny{$\pm$4.0} & 43.5\tiny{$\pm$4.7} & 99.9\tiny{$\pm$0.0} & 61.6\tiny{$\pm$4.0} & 44.6\tiny{$\pm$4.6} & 99.8\tiny{$\pm$0.1} \\
& DER++ & 72.4\tiny{$\pm$4.4} & 32.2\tiny{$\pm$5.1} & 99.9\tiny{$\pm$0.0} & 75.1\tiny{$\pm$4.1} & 29.0\tiny{$\pm$4.8} & 99.9\tiny{$\pm$0.0} & 70.7\tiny{$\pm$2.9} & 34.0\tiny{$\pm$3.3} & 99.8\tiny{$\pm$0.1} \\
& RanPAC & 81.9\tiny{$\pm$0.3} & 10.0\tiny{$\pm$1.9} & 90.4\tiny{$\pm$1.6} & 83.1\tiny{$\pm$0.1} & 8.0\tiny{$\pm$1.5} & 89.9\tiny{$\pm$1.3} & 82.1\tiny{$\pm$1.3} & 11.6\tiny{$\pm$2.5} & 92.1\tiny{$\pm$2.1} \\
& FeCAM & 55.3\tiny{$\pm$1.3} & 6.8\tiny{$\pm$1.8} & 61.1\tiny{$\pm$2.5} & 60.1\tiny{$\pm$0.1} & 3.5\tiny{$\pm$0.7} & 63.1\tiny{$\pm$0.7} & 56.6\tiny{$\pm$1.0} & 5.8\tiny{$\pm$1.7} & 61.5\tiny{$\pm$0.7} \\
\bottomrule
\end{tabular}
}
\end{table*}

\begin{table*}[h]
\centering
\scriptsize
\setlength{\tabcolsep}{3pt}
\renewcommand{\arraystretch}{1.0}
\caption{Per-encoder breakdown of the protocol sanity check (part 2/2: GearNet-Edge, ProtBert, SaProt). Same format as Table~\ref{tab:baseline_protocol_sanity_per_encoder_a}. Values are mean$\pm$std over three seeds (\%). Acc = Avg Acc ($\uparrow$), Fgt = Forgetting ($\downarrow$), Plast = Plasticity ($\uparrow$).}
\label{tab:baseline_protocol_sanity_per_encoder_b}
\vspace{2pt}
\resizebox{\textwidth}{!}{%
\begin{tabular}{ll | ccc | ccc | ccc}
\toprule
\multirow{2}{*}{\textbf{Dataset}}
& \multirow{2}{*}{\textbf{Method}}
& \multicolumn{3}{c|}{\textbf{Random classes}}
& \multicolumn{3}{c|}{\textbf{Finer-EC buckets}}
& \multicolumn{3}{c}{\textbf{Mixed hierarchy}} \\
\cmidrule(lr){3-5} \cmidrule(lr){6-8} \cmidrule(lr){9-11}
& 
& \textbf{Acc} $\uparrow$ & \textbf{Fgt} $\downarrow$ & \textbf{Plast} $\uparrow$
& \textbf{Acc} $\uparrow$ & \textbf{Fgt} $\downarrow$ & \textbf{Plast} $\uparrow$
& \textbf{Acc} $\uparrow$ & \textbf{Fgt} $\downarrow$ & \textbf{Plast} $\uparrow$ \\
\midrule
\multicolumn{11}{c}{\textbf{\textit{GearNet-Edge}}} \\
\midrule
\multirow{9}{*}{\shortstack[l]{\textbf{Enzyme}\\\textbf{Reaction}}}
& EWC & 10.5\tiny{$\pm$0.7} & 68.3\tiny{$\pm$2.0} & 69.1\tiny{$\pm$2.4} & 9.9\tiny{$\pm$1.2} & 65.0\tiny{$\pm$1.8} & 65.5\tiny{$\pm$1.3} & 10.0\tiny{$\pm$0.3} & 70.7\tiny{$\pm$2.0} & 70.6\tiny{$\pm$1.9} \\
& FeTrIL & 17.5\tiny{$\pm$3.9} & 57.3\tiny{$\pm$4.1} & 66.6\tiny{$\pm$2.8} & 18.5\tiny{$\pm$2.9} & 54.3\tiny{$\pm$4.7} & 64.6\tiny{$\pm$1.5} & 17.7\tiny{$\pm$4.8} & 58.6\tiny{$\pm$4.6} & 67.9\tiny{$\pm$0.9} \\
& EASE & 24.0\tiny{$\pm$3.2} & 15.0\tiny{$\pm$4.1} & 36.9\tiny{$\pm$1.1} & 27.1\tiny{$\pm$0.9} & 12.8\tiny{$\pm$2.8} & 37.8\tiny{$\pm$1.9} & 25.0\tiny{$\pm$2.6} & 13.3\tiny{$\pm$3.3} & 36.4\tiny{$\pm$0.4} \\
& RER & 36.6\tiny{$\pm$2.8} & 33.6\tiny{$\pm$2.2} & 64.7\tiny{$\pm$2.8} & 36.3\tiny{$\pm$1.4} & 31.2\tiny{$\pm$2.1} & 62.3\tiny{$\pm$2.1} & 37.5\tiny{$\pm$2.1} & 32.2\tiny{$\pm$2.7} & 65.1\tiny{$\pm$0.3} \\
& Replay & 36.5\tiny{$\pm$1.9} & 34.9\tiny{$\pm$2.5} & 65.7\tiny{$\pm$2.9} & 35.9\tiny{$\pm$1.2} & 32.7\tiny{$\pm$0.5} & 63.3\tiny{$\pm$1.9} & 37.9\tiny{$\pm$2.1} & 33.0\tiny{$\pm$3.1} & 66.2\tiny{$\pm$0.5} \\
& iCaRL & 36.8\tiny{$\pm$2.5} & 33.4\tiny{$\pm$1.7} & 65.2\tiny{$\pm$2.0} & 40.2\tiny{$\pm$2.7} & 28.8\tiny{$\pm$2.8} & 64.0\tiny{$\pm$2.6} & 40.9\tiny{$\pm$3.8} & 29.9\tiny{$\pm$3.9} & 66.6\tiny{$\pm$0.4} \\
& DER++ & 37.8\tiny{$\pm$3.1} & 29.5\tiny{$\pm$3.5} & 62.0\tiny{$\pm$3.4} & 39.8\tiny{$\pm$1.2} & 25.3\tiny{$\pm$2.5} & 60.2\tiny{$\pm$0.6} & 40.3\tiny{$\pm$2.0} & 28.0\tiny{$\pm$2.3} & 64.0\tiny{$\pm$2.2} \\
& RanPAC & 24.6\tiny{$\pm$2.2} & 13.4\tiny{$\pm$5.4} & 35.6\tiny{$\pm$6.8} & 24.0\tiny{$\pm$0.2} & 9.3\tiny{$\pm$4.6} & 31.5\tiny{$\pm$4.1} & 24.8\tiny{$\pm$0.6} & 10.5\tiny{$\pm$2.0} & 33.7\tiny{$\pm$1.2} \\
& FeCAM & 39.8\tiny{$\pm$3.9} & 11.1\tiny{$\pm$3.3} & 49.4\tiny{$\pm$3.4} & 40.4\tiny{$\pm$1.4} & 9.9\tiny{$\pm$3.1} & 48.8\tiny{$\pm$1.3} & 39.7\tiny{$\pm$3.0} & 9.9\tiny{$\pm$3.6} & 48.1\tiny{$\pm$0.8} \\
\cmidrule(l){1-11}
\multirow{9}{*}{\shortstack[l]{\textbf{EC}\\\textbf{Number}}}
& EWC & 15.7\tiny{$\pm$1.0} & 90.4\tiny{$\pm$3.8} & 93.2\tiny{$\pm$3.4} & 15.5\tiny{$\pm$0.9} & 90.8\tiny{$\pm$2.1} & 93.4\tiny{$\pm$2.6} & 16.4\tiny{$\pm$1.9} & 85.3\tiny{$\pm$5.6} & 89.5\tiny{$\pm$6.6} \\
& FeTrIL & 20.5\tiny{$\pm$6.5} & 85.0\tiny{$\pm$8.3} & 93.1\tiny{$\pm$1.2} & 17.0\tiny{$\pm$5.1} & 89.0\tiny{$\pm$6.6} & 93.3\tiny{$\pm$0.6} & 22.8\tiny{$\pm$9.2} & 81.3\tiny{$\pm$12.4} & 91.0\tiny{$\pm$2.2} \\
& EASE & 44.4\tiny{$\pm$0.9} & 6.5\tiny{$\pm$1.1} & 49.9\tiny{$\pm$0.7} & 48.6\tiny{$\pm$0.1} & 4.2\tiny{$\pm$1.9} & 52.0\tiny{$\pm$1.7} & 46.3\tiny{$\pm$1.3} & 4.4\tiny{$\pm$2.2} & 49.7\tiny{$\pm$2.5} \\
& RER & 34.5\tiny{$\pm$0.3} & 71.8\tiny{$\pm$3.4} & 94.0\tiny{$\pm$2.8} & 34.8\tiny{$\pm$1.2} & 71.2\tiny{$\pm$1.2} & 94.3\tiny{$\pm$2.5} & 35.1\tiny{$\pm$1.0} & 69.5\tiny{$\pm$3.3} & 91.2\tiny{$\pm$5.4} \\
& Replay & 34.7\tiny{$\pm$0.4} & 71.7\tiny{$\pm$3.3} & 94.1\tiny{$\pm$2.7} & 34.4\tiny{$\pm$0.9} & 71.6\tiny{$\pm$1.4} & 94.3\tiny{$\pm$2.3} & 34.5\tiny{$\pm$1.0} & 70.1\tiny{$\pm$4.1} & 91.3\tiny{$\pm$5.2} \\
& iCaRL & 37.4\tiny{$\pm$3.2} & 69.6\tiny{$\pm$2.8} & 95.6\tiny{$\pm$2.6} & 36.8\tiny{$\pm$3.0} & 70.5\tiny{$\pm$2.1} & 95.7\tiny{$\pm$1.9} & 34.7\tiny{$\pm$3.3} & 72.3\tiny{$\pm$2.1} & 93.4\tiny{$\pm$4.1} \\
& DER++ & 41.7\tiny{$\pm$3.4} & 62.2\tiny{$\pm$6.6} & 93.0\tiny{$\pm$2.6} & 41.7\tiny{$\pm$3.6} & 61.0\tiny{$\pm$6.5} & 92.7\tiny{$\pm$2.8} & 40.2\tiny{$\pm$4.0} & 63.5\tiny{$\pm$9.1} & 90.4\tiny{$\pm$5.5} \\
& RanPAC & 22.4\tiny{$\pm$1.1} & 6.3\tiny{$\pm$2.9} & 27.8\tiny{$\pm$2.6} & 19.7\tiny{$\pm$0.2} & 4.8\tiny{$\pm$2.5} & 23.8\tiny{$\pm$1.9} & 22.5\tiny{$\pm$3.1} & 7.8\tiny{$\pm$2.9} & 29.1\tiny{$\pm$3.4} \\
& FeCAM & 39.1\tiny{$\pm$0.6} & 6.5\tiny{$\pm$1.7} & 44.6\tiny{$\pm$1.8} & 45.0\tiny{$\pm$0.2} & 4.2\tiny{$\pm$1.4} & 48.6\tiny{$\pm$1.3} & 39.8\tiny{$\pm$1.4} & 6.1\tiny{$\pm$1.6} & 45.0\tiny{$\pm$1.1} \\
\midrule
\multicolumn{11}{c}{\textbf{\textit{ProtBert}}} \\
\midrule
\multirow{9}{*}{\shortstack[l]{\textbf{Enzyme}\\\textbf{Reaction}}}
& EWC & 13.2\tiny{$\pm$0.8} & 76.9\tiny{$\pm$2.6} & 79.1\tiny{$\pm$2.9} & 13.9\tiny{$\pm$1.8} & 70.6\tiny{$\pm$1.4} & 74.4\tiny{$\pm$0.9} & 15.3\tiny{$\pm$2.0} & 76.0\tiny{$\pm$1.9} & 80.5\tiny{$\pm$1.1} \\
& FeTrIL & 21.3\tiny{$\pm$5.1} & 69.0\tiny{$\pm$6.3} & 80.5\tiny{$\pm$2.0} & 22.6\tiny{$\pm$4.8} & 64.2\tiny{$\pm$5.5} & 77.7\tiny{$\pm$0.2} & 21.5\tiny{$\pm$5.3} & 68.9\tiny{$\pm$6.1} & 80.6\tiny{$\pm$0.7} \\
& EASE & 37.1\tiny{$\pm$1.8} & 14.3\tiny{$\pm$3.3} & 49.2\tiny{$\pm$1.2} & 40.3\tiny{$\pm$0.2} & 13.0\tiny{$\pm$2.5} & 51.3\tiny{$\pm$2.4} & 36.1\tiny{$\pm$4.1} & 14.8\tiny{$\pm$2.4} & 48.3\tiny{$\pm$1.8} \\
& RER & 43.5\tiny{$\pm$2.4} & 38.0\tiny{$\pm$2.4} & 76.0\tiny{$\pm$2.8} & 46.6\tiny{$\pm$1.4} & 30.2\tiny{$\pm$1.9} & 72.3\tiny{$\pm$1.6} & 47.7\tiny{$\pm$3.4} & 35.1\tiny{$\pm$3.7} & 77.8\tiny{$\pm$0.9} \\
& Replay & 43.8\tiny{$\pm$3.0} & 37.9\tiny{$\pm$2.1} & 76.3\tiny{$\pm$2.7} & 45.9\tiny{$\pm$1.4} & 31.2\tiny{$\pm$1.2} & 72.5\tiny{$\pm$0.8} & 47.9\tiny{$\pm$3.5} & 34.3\tiny{$\pm$3.1} & 77.3\tiny{$\pm$1.2} \\
& iCaRL & 48.5\tiny{$\pm$2.6} & 31.7\tiny{$\pm$1.6} & 75.6\tiny{$\pm$1.3} & 50.2\tiny{$\pm$1.5} & 25.7\tiny{$\pm$1.2} & 71.9\tiny{$\pm$0.7} & 51.3\tiny{$\pm$3.2} & 29.5\tiny{$\pm$2.7} & 76.6\tiny{$\pm$0.9} \\
& DER++ & 55.1\tiny{$\pm$4.0} & 19.4\tiny{$\pm$3.1} & 71.5\tiny{$\pm$4.0} & 57.4\tiny{$\pm$5.6} & 16.1\tiny{$\pm$3.7} & 69.9\tiny{$\pm$3.2} & 57.5\tiny{$\pm$3.2} & 18.8\tiny{$\pm$1.9} & 73.2\tiny{$\pm$2.6} \\
& RanPAC & 65.2\tiny{$\pm$3.2} & 12.4\tiny{$\pm$2.1} & 75.8\tiny{$\pm$3.3} & 67.4\tiny{$\pm$0.2} & 6.3\tiny{$\pm$1.5} & 72.6\tiny{$\pm$1.3} & 65.6\tiny{$\pm$1.6} & 10.9\tiny{$\pm$1.5} & 74.9\tiny{$\pm$0.4} \\
& FeCAM & 65.7\tiny{$\pm$4.7} & 10.0\tiny{$\pm$3.9} & 74.3\tiny{$\pm$1.7} & 67.7\tiny{$\pm$0.2} & 6.9\tiny{$\pm$0.9} & 73.6\tiny{$\pm$0.4} & 65.9\tiny{$\pm$3.1} & 11.0\tiny{$\pm$3.5} & 75.3\tiny{$\pm$0.6} \\
\cmidrule(l){1-11}
\multirow{9}{*}{\shortstack[l]{\textbf{EC}\\\textbf{Number}}}
& EWC & 36.9\tiny{$\pm$4.1} & 73.5\tiny{$\pm$4.8} & 99.9\tiny{$\pm$0.0} & 43.1\tiny{$\pm$6.4} & 66.2\tiny{$\pm$7.5} & 99.8\tiny{$\pm$0.0} & 31.0\tiny{$\pm$5.8} & 80.4\tiny{$\pm$6.8} & 99.9\tiny{$\pm$0.0} \\
& FeTrIL & 23.2\tiny{$\pm$7.5} & 89.5\tiny{$\pm$8.8} & 99.9\tiny{$\pm$0.0} & 19.1\tiny{$\pm$4.6} & 94.2\tiny{$\pm$5.4} & 99.8\tiny{$\pm$0.0} & 25.4\tiny{$\pm$9.8} & 86.8\tiny{$\pm$11.4} & 99.8\tiny{$\pm$0.1} \\
& EASE & 56.5\tiny{$\pm$0.4} & 10.0\tiny{$\pm$1.5} & 64.9\tiny{$\pm$1.6} & 52.7\tiny{$\pm$0.7} & 7.2\tiny{$\pm$4.8} & 58.7\tiny{$\pm$4.5} & 56.9\tiny{$\pm$1.5} & 8.1\tiny{$\pm$3.6} & 63.3\tiny{$\pm$3.5} \\
& RER & 52.0\tiny{$\pm$3.6} & 55.8\tiny{$\pm$4.3} & 99.9\tiny{$\pm$0.1} & 56.0\tiny{$\pm$3.9} & 51.2\tiny{$\pm$4.5} & 99.9\tiny{$\pm$0.0} & 50.6\tiny{$\pm$2.4} & 57.4\tiny{$\pm$2.9} & 99.8\tiny{$\pm$0.1} \\
& Replay & 52.8\tiny{$\pm$2.8} & 55.0\tiny{$\pm$3.3} & 99.9\tiny{$\pm$0.0} & 56.8\tiny{$\pm$3.7} & 50.2\tiny{$\pm$4.3} & 99.9\tiny{$\pm$0.0} & 50.6\tiny{$\pm$2.5} & 57.5\tiny{$\pm$2.8} & 99.8\tiny{$\pm$0.1} \\
& iCaRL & 58.3\tiny{$\pm$5.5} & 48.5\tiny{$\pm$6.5} & 99.9\tiny{$\pm$0.0} & 57.5\tiny{$\pm$3.4} & 49.4\tiny{$\pm$4.0} & 99.9\tiny{$\pm$0.0} & 55.2\tiny{$\pm$2.3} & 52.0\tiny{$\pm$2.6} & 99.9\tiny{$\pm$0.0} \\
& DER++ & 68.1\tiny{$\pm$5.9} & 37.2\tiny{$\pm$6.9} & 99.9\tiny{$\pm$0.0} & 69.3\tiny{$\pm$4.4} & 35.7\tiny{$\pm$5.1} & 99.9\tiny{$\pm$0.0} & 67.3\tiny{$\pm$3.5} & 38.0\tiny{$\pm$4.1} & 99.9\tiny{$\pm$0.0} \\
& RanPAC & 76.6\tiny{$\pm$0.8} & 11.5\tiny{$\pm$1.9} & 86.4\tiny{$\pm$2.4} & 78.0\tiny{$\pm$0.1} & 8.4\tiny{$\pm$2.0} & 85.2\tiny{$\pm$1.6} & 76.6\tiny{$\pm$1.4} & 13.9\tiny{$\pm$3.5} & 88.6\tiny{$\pm$2.9} \\
& FeCAM & 30.1\tiny{$\pm$0.4} & 5.1\tiny{$\pm$1.8} & 34.4\tiny{$\pm$1.7} & 34.0\tiny{$\pm$0.1} & 3.1\tiny{$\pm$0.4} & 36.7\tiny{$\pm$0.5} & 31.2\tiny{$\pm$1.6} & 4.1\tiny{$\pm$1.4} & 34.8\tiny{$\pm$1.0} \\
\midrule
\multicolumn{11}{c}{\textbf{\textit{SaProt}}} \\
\midrule
\multirow{9}{*}{\shortstack[l]{\textbf{Enzyme}\\\textbf{Reaction}}}
& EWC & 17.9\tiny{$\pm$2.2} & 77.9\tiny{$\pm$2.8} & 84.7\tiny{$\pm$0.3} & 17.4\tiny{$\pm$1.2} & 73.5\tiny{$\pm$1.2} & 80.4\tiny{$\pm$0.2} & 17.5\tiny{$\pm$1.8} & 78.0\tiny{$\pm$2.5} & 84.4\tiny{$\pm$1.0} \\
& FeTrIL & 22.7\tiny{$\pm$4.6} & 71.0\tiny{$\pm$5.7} & 83.6\tiny{$\pm$0.5} & 25.6\tiny{$\pm$5.9} & 65.2\tiny{$\pm$6.5} & 81.5\tiny{$\pm$0.6} & 24.2\tiny{$\pm$5.5} & 70.0\tiny{$\pm$5.7} & 84.3\tiny{$\pm$0.7} \\
& EASE & 46.1\tiny{$\pm$2.2} & 15.4\tiny{$\pm$4.8} & 59.2\tiny{$\pm$2.8} & 51.8\tiny{$\pm$1.8} & 13.1\tiny{$\pm$2.6} & 62.8\tiny{$\pm$2.2} & 47.4\tiny{$\pm$3.1} & 14.1\tiny{$\pm$4.0} & 59.2\tiny{$\pm$0.2} \\
& RER & 52.2\tiny{$\pm$3.7} & 34.3\tiny{$\pm$5.2} & 81.6\tiny{$\pm$0.7} & 54.8\tiny{$\pm$1.9} & 28.6\tiny{$\pm$2.2} & 79.3\tiny{$\pm$0.1} & 53.6\tiny{$\pm$1.8} & 32.5\tiny{$\pm$2.2} & 81.5\tiny{$\pm$1.1} \\
& Replay & 51.9\tiny{$\pm$3.2} & 34.7\tiny{$\pm$4.5} & 81.7\tiny{$\pm$0.6} & 54.6\tiny{$\pm$2.2} & 28.6\tiny{$\pm$2.4} & 79.1\tiny{$\pm$0.1} & 53.4\tiny{$\pm$1.4} & 33.4\tiny{$\pm$1.7} & 82.0\tiny{$\pm$0.9} \\
& iCaRL & 54.2\tiny{$\pm$4.3} & 29.0\tiny{$\pm$4.8} & 79.0\tiny{$\pm$1.4} & 60.6\tiny{$\pm$0.6} & 20.9\tiny{$\pm$0.4} & 78.5\tiny{$\pm$0.8} & 61.0\tiny{$\pm$1.8} & 22.9\tiny{$\pm$3.4} & 80.6\tiny{$\pm$1.4} \\
& DER++ & 60.7\tiny{$\pm$2.9} & 16.5\tiny{$\pm$1.2} & 74.6\tiny{$\pm$3.1} & 64.3\tiny{$\pm$5.0} & 14.1\tiny{$\pm$2.9} & 75.7\tiny{$\pm$2.9} & 62.7\tiny{$\pm$3.4} & 15.2\tiny{$\pm$0.6} & 75.7\tiny{$\pm$3.5} \\
& RanPAC & 69.7\tiny{$\pm$3.7} & 12.6\tiny{$\pm$3.5} & 80.2\tiny{$\pm$2.4} & 75.4\tiny{$\pm$0.3} & 3.6\tiny{$\pm$0.6} & 78.5\tiny{$\pm$0.4} & 71.5\tiny{$\pm$1.1} & 9.3\tiny{$\pm$0.9} & 79.5\tiny{$\pm$0.5} \\
& FeCAM & 65.9\tiny{$\pm$4.0} & 12.8\tiny{$\pm$6.0} & 76.9\tiny{$\pm$1.9} & 68.3\tiny{$\pm$0.8} & 7.7\tiny{$\pm$1.3} & 74.8\tiny{$\pm$0.6} & 66.6\tiny{$\pm$2.3} & 11.0\tiny{$\pm$3.3} & 76.1\tiny{$\pm$0.6} \\
\cmidrule(l){1-11}
\multirow{9}{*}{\shortstack[l]{\textbf{EC}\\\textbf{Number}}}
& EWC & 34.3\tiny{$\pm$4.7} & 76.5\tiny{$\pm$5.4} & 99.9\tiny{$\pm$0.0} & 38.3\tiny{$\pm$7.7} & 71.7\tiny{$\pm$8.8} & 99.8\tiny{$\pm$0.1} & 33.5\tiny{$\pm$5.2} & 77.5\tiny{$\pm$6.0} & 99.9\tiny{$\pm$0.0} \\
& FeTrIL & 23.1\tiny{$\pm$7.6} & 89.6\tiny{$\pm$8.9} & 99.9\tiny{$\pm$0.0} & 19.3\tiny{$\pm$4.8} & 93.9\tiny{$\pm$5.5} & 99.8\tiny{$\pm$0.1} & 25.8\tiny{$\pm$9.9} & 86.4\tiny{$\pm$11.6} & 99.9\tiny{$\pm$0.0} \\
& EASE & 59.0\tiny{$\pm$1.1} & 8.0\tiny{$\pm$0.5} & 65.9\tiny{$\pm$1.5} & 59.0\tiny{$\pm$0.3} & 5.6\tiny{$\pm$3.6} & 63.4\tiny{$\pm$3.2} & 60.7\tiny{$\pm$0.9} & 6.7\tiny{$\pm$2.3} & 66.2\tiny{$\pm$2.0} \\
& RER & 51.1\tiny{$\pm$2.3} & 56.9\tiny{$\pm$2.6} & 99.9\tiny{$\pm$0.0} & 56.6\tiny{$\pm$3.0} & 50.3\tiny{$\pm$3.4} & 99.8\tiny{$\pm$0.1} & 46.6\tiny{$\pm$3.9} & 62.0\tiny{$\pm$4.5} & 99.8\tiny{$\pm$0.1} \\
& Replay & 50.1\tiny{$\pm$1.5} & 58.2\tiny{$\pm$1.7} & 99.9\tiny{$\pm$0.0} & 54.7\tiny{$\pm$4.4} & 52.6\tiny{$\pm$5.1} & 99.8\tiny{$\pm$0.1} & 47.7\tiny{$\pm$2.4} & 60.9\tiny{$\pm$2.8} & 99.9\tiny{$\pm$0.0} \\
& iCaRL & 56.7\tiny{$\pm$3.5} & 50.5\tiny{$\pm$4.0} & 99.9\tiny{$\pm$0.0} & 61.6\tiny{$\pm$1.9} & 44.6\tiny{$\pm$2.1} & 99.9\tiny{$\pm$0.0} & 57.4\tiny{$\pm$5.0} & 49.6\tiny{$\pm$5.8} & 99.9\tiny{$\pm$0.0} \\
& DER++ & 67.2\tiny{$\pm$5.6} & 38.1\tiny{$\pm$6.6} & 99.9\tiny{$\pm$0.0} & 69.4\tiny{$\pm$4.2} & 35.6\tiny{$\pm$4.9} & 99.9\tiny{$\pm$0.0} & 65.8\tiny{$\pm$3.0} & 39.8\tiny{$\pm$3.5} & 99.9\tiny{$\pm$0.0} \\
& RanPAC & 83.8\tiny{$\pm$0.7} & 8.2\tiny{$\pm$1.1} & 90.8\tiny{$\pm$1.6} & 85.1\tiny{$\pm$0.1} & 5.7\tiny{$\pm$1.2} & 89.9\tiny{$\pm$1.0} & 83.8\tiny{$\pm$0.7} & 9.6\tiny{$\pm$1.9} & 92.0\tiny{$\pm$1.8} \\
& FeCAM & 47.5\tiny{$\pm$1.2} & 6.9\tiny{$\pm$2.6} & 53.4\tiny{$\pm$2.8} & 52.0\tiny{$\pm$0.1} & 4.0\tiny{$\pm$1.5} & 55.4\tiny{$\pm$1.4} & 49.5\tiny{$\pm$1.5} & 5.4\tiny{$\pm$1.6} & 54.1\tiny{$\pm$1.6} \\
\bottomrule
\end{tabular}
}
\end{table*}

\clearpage
\subsection{Cross-family confusion patterns.}
Across both datasets, the dominant cross-family errors consistently involve transferases and hydrolases, indicating that these families remain difficult to separate under a single shared covariance. However, the magnitude of these confusions varies substantially across encoders and datasets. This variation further supports the conclusion that the effectiveness of a shared covariance assumption depends strongly on the geometry induced by the underlying protein representation.

\begin{table*}[h]
\centering
\caption{Cross-family confusion breakdowns under a single shared covariance for all five protein encoders on both datasets. For each encoder and dataset, we report the six most frequent cross-family misclassifications together with one representative EC-level confusion.}
\label{tab:full_confusion}
\scriptsize
\setlength{\tabcolsep}{3.5pt}
\renewcommand{\arraystretch}{0.95}
\resizebox{\textwidth}{!}{%
\begin{tabular}{llrl|llrl}
\toprule
\multicolumn{4}{c|}{\textbf{Enzyme Reaction}}
& \multicolumn{4}{c}{\textbf{EC Number}} \\
True & Pred.\ & \# mis. & Example
& True & Pred.\ & \# mis. & Example \\
\midrule

\multicolumn{4}{c|}{\textbf{ESM2-150M}}
& \multicolumn{4}{c}{\textbf{ESM2-150M}} \\
\midrule
2 (transf.) & 3 (hydro.) & 96 & 2.1.1.- $\to$ 3.1.21.4
& 3 (hydro.) & 2 (transf.) & 461 & 3.1.4.35 $\to$ 2.4.1.229 \\
3 (hydro.) & 2 (transf.) & 48 & 3.4.19.12 $\to$ 2.7.10.-
& 2 (transf.) & 3 (hydro.) & 270 & 2.7.2.3 $\to$ 3.5.1.76 \\
2 (transf.) & 4 (lyases) & 34 & 2.7.7.- $\to$ 4.2.1.-
& 2 (transf.) & 1 (oxidored.) & 170 & 2.7.1.148 $\to$ 1.1.1.193 \\
3 (hydro.) & 5 (isomer.) & 16 & 3.6.4.- $\to$ 5.6.1.-
& 3 (hydro.) & 1 (oxidored.) & 159 & 3.1.21.4 $\to$ 1.13.11.58 \\
2 (transf.) & 1 (oxidored.) & 15 & 2.7.11.1 $\to$ 1.13.12.-
& 2 (transf.) & 4 (lyases) & 117 & 2.4.2.9 $\to$ 4.2.1.91 \\
4 (lyases) & 2 (transf.) & 14 & 4.1.1.- $\to$ 2.7.4.6
& 2 (transf.) & 6 (ligases) & 102 & 2.7.7.101 $\to$ 6.5.1.2 \\

\midrule
\multicolumn{4}{c|}{\textbf{ESM2-650M}}
& \multicolumn{4}{c}{\textbf{ESM2-650M}} \\
\midrule
2 (transf.) & 3 (hydro.) & 59 & 2.7.7.- $\to$ 3.1.13.-
& 3 (hydro.) & 2 (transf.) & 425 & 3.1.3.16 $\to$ 2.3.1.84 \\
3 (hydro.) & 2 (transf.) & 53 & 3.4.22.28 $\to$ 2.5.1.-
& 2 (transf.) & 3 (hydro.) & 270 & 2.3.1.48 $\to$ 3.1.21.4 \\
2 (transf.) & 4 (lyases) & 31 & 2.7.7.48 $\to$ 4.2.1.-
& 2 (transf.) & 1 (oxidored.) & 159 & 2.7.1.148 $\to$ 1.1.1.193 \\
3 (hydro.) & 4 (lyases) & 21 & 3.6.4.- $\to$ 4.2.1.-
& 3 (hydro.) & 1 (oxidored.) & 124 & 3.1.4.17 $\to$ 1.14.11.30 \\
2 (transf.) & 6 (ligases) & 11 & 2.4.2.- $\to$ 6.3.5.2
& 2 (transf.) & 6 (ligases) & 124 & 2.7.2.3 $\to$ 6.4.1.2 \\
4 (lyases) & 5 (isomer.) & 9 & 4.2.1.- $\to$ 5.3.3.8
& 2 (transf.) & 4 (lyases) & 92 & 2.3.2.27 $\to$ 4.99.1.10 \\

\midrule
\multicolumn{4}{c|}{\textbf{GearNet-Edge}}
& \multicolumn{4}{c}{\textbf{GearNet-Edge}} \\
\midrule
2 (transf.) & 3 (hydro.) & 124 & 2.4.2.- $\to$ 3.1.13.-
& 3 (hydro.) & 2 (transf.) & 369 & 3.2.1.15 $\to$ 2.7.1.161 \\
2 (transf.) & 1 (oxidored.) & 78 & 2.5.1.- $\to$ 1.2.1.11
& 2 (transf.) & 3 (hydro.) & 277 & 2.3.2.27 $\to$ 3.6.1.62 \\
3 (hydro.) & 2 (transf.) & 71 & 3.2.1.4 $\to$ 2.4.1.-
& 2 (transf.) & 1 (oxidored.) & 112 & 2.8.4.1 $\to$ 1.1.99.37 \\
3 (hydro.) & 1 (oxidored.) & 55 & 3.2.1.4 $\to$ 1.8.1.-
& 3 (hydro.) & 1 (oxidored.) & 105 & 3.5.1.28 $\to$ 1.6.99.1 \\
2 (transf.) & 6 (ligases) & 39 & 2.7.1.- $\to$ 6.1.1.15
& 2 (transf.) & 4 (lyases) & 73 & 2.3.1.15 $\to$ 4.1.1.31 \\
2 (transf.) & 4 (lyases) & 37 & 2.7.11.1 $\to$ 4.2.99.-
& 1 (oxidored.) & 2 (transf.) & 71 & 1.15.1.1 $\to$ 2.4.2.31 \\

\midrule
\multicolumn{4}{c|}{\textbf{ProtBert}}
& \multicolumn{4}{c}{\textbf{ProtBert}} \\
\midrule
2 (transf.) & 3 (hydro.) & 106 & 2.7.11.1 $\to$ 3.4.19.12
& 3 (hydro.) & 2 (transf.) & 124 & 3.1.4.35 $\to$ 2.7.11.1 \\
3 (hydro.) & 2 (transf.) & 59 & 3.6.4.- $\to$ 2.7.7.-
& 2 (transf.) & 3 (hydro.) & 123 & 2.7.11.1 $\to$ 3.4.19.12 \\
3 (hydro.) & 4 (lyases) & 21 & 3.5.1.2 $\to$ 4.2.1.-
& 1 (oxidored.) & 3 (hydro.) & 37 & 1.15.1.1 $\to$ 3.1.1.3 \\
2 (transf.) & 4 (lyases) & 12 & 2.7.1.- $\to$ 4.2.1.-
& 3 (hydro.) & 1 (oxidored.) & 34 & 3.1.1.4 $\to$ 1.4.3.2 \\
4 (lyases) & 2 (transf.) & 12 & 4.1.1.- $\to$ 2.7.7.48
& 7 (transl.) & 1 (oxidored.) & 31 & 7.1.1.9 $\to$ 1.12.98.3 \\
2 (transf.) & 1 (oxidored.) & 11 & 2.7.1.- $\to$ 1.16.3.1
& 4 (lyases) & 3 (hydro.) & 24 & 4.2.2.10 $\to$ 3.2.1.133 \\

\midrule
\multicolumn{4}{c|}{\textbf{SaProt}}
& \multicolumn{4}{c}{\textbf{SaProt}} \\
\midrule
2 (transf.) & 3 (hydro.) & 89 & 2.7.7.- $\to$ 3.1.13.-
& 2 (transf.) & 3 (hydro.) & 84 & 2.3.2.27 $\to$ 3.2.2.6 \\
3 (hydro.) & 2 (transf.) & 64 & 3.1.3.- $\to$ 2.8.1.1
& 3 (hydro.) & 2 (transf.) & 77 & 3.5.4.26 $\to$ 2.1.1.85 \\
2 (transf.) & 6 (ligases) & 18 & 2.4.2.- $\to$ 6.3.5.2
& 3 (hydro.) & 1 (oxidored.) & 39 & 3.5.4.19 $\to$ 1.1.1.23 \\
2 (transf.) & 4 (lyases) & 15 & 2.7.11.1 $\to$ 4.6.1.2
& 7 (transl.) & 1 (oxidored.) & 33 & 7.1.1.9 $\to$ 1.7.2.5 \\
2 (transf.) & 1 (oxidored.) & 14 & 2.7.13.- $\to$ 1.6.5.-
& 2 (transf.) & 1 (oxidored.) & 28 & 2.7.2.3 $\to$ 1.2.1.28 \\
4 (lyases) & 2 (transf.) & 12 & 4.1.1.- $\to$ 2.7.7.6
& 2 (transf.) & 4 (lyases) & 22 & 2.5.1.18 $\to$ 4.4.1.20 \\

\bottomrule
\end{tabular}%
}
\end{table*}


\clearpage

\begin{figure*}[h]
\centering
\includegraphics[width=\textwidth]{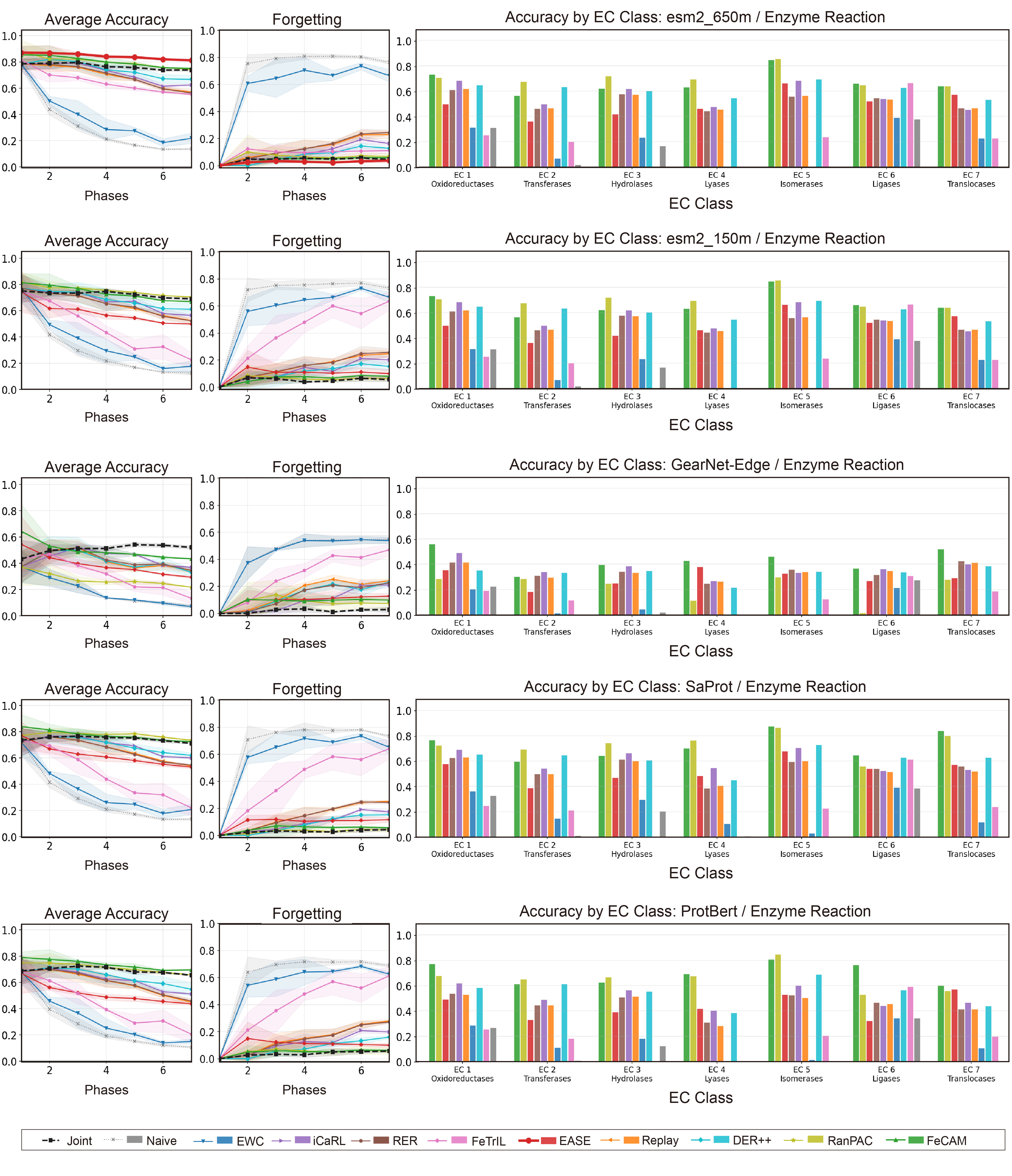}
\caption{Per-encoder continual-learning results on Enzyme Reaction (348 classes, 7 reaction type groups). \textbf{Left:} Average incremental accuracy across phases. \textbf{Middle:} Forgetting across phases. \textbf{Right:} Final accuracy broken down by reaction type.}
\label{fig:enzyme_reaction_curves}
\end{figure*}

\begin{figure*}[h]
\centering
\includegraphics[width=\textwidth]{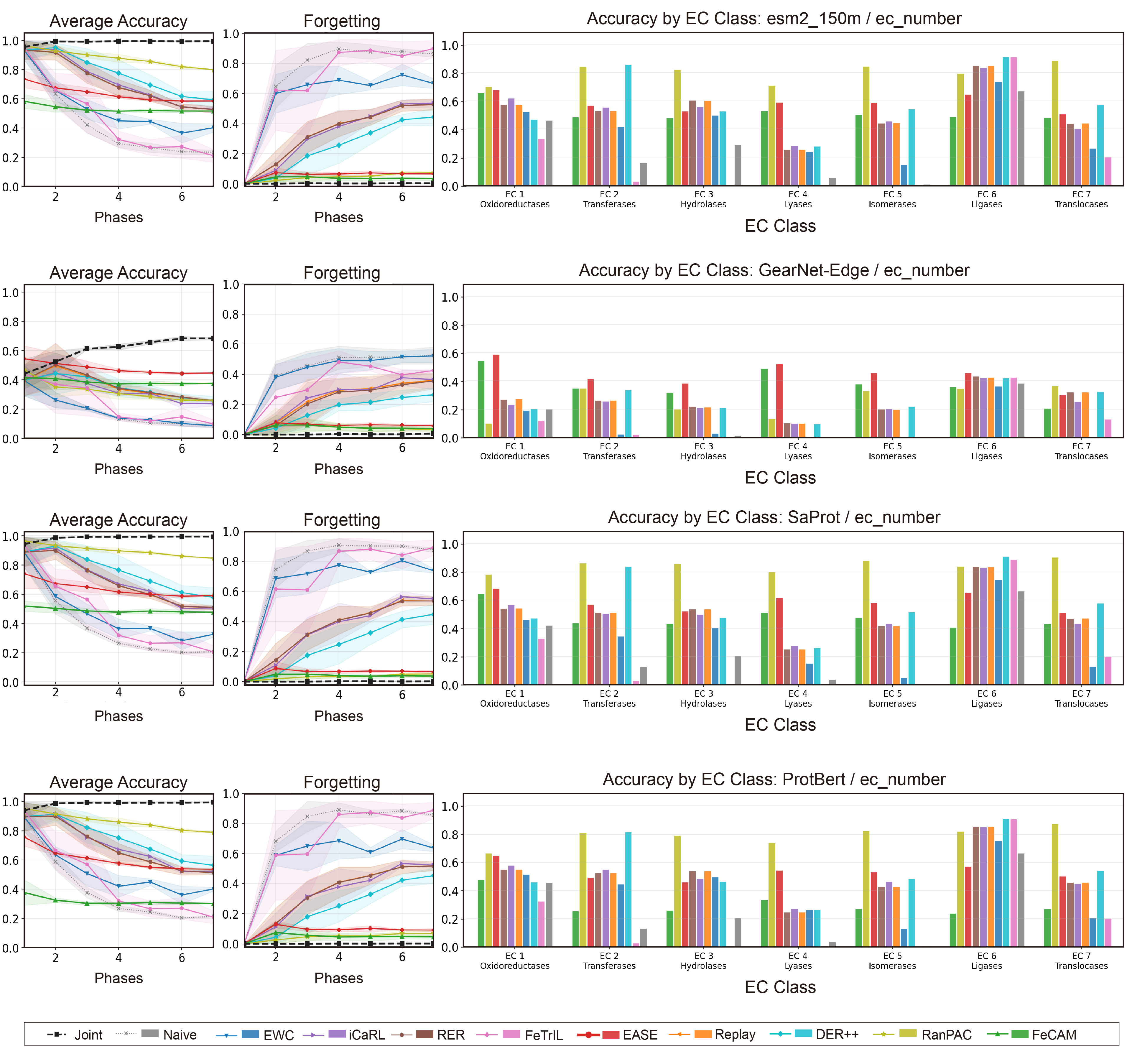}
\caption{Per-encoder continual-learning results on EC Number (4{,}788 classes, 7 EC top-level groups). \textbf{Left:} Average incremental accuracy across phases. \textbf{Middle:} Forgetting across phases. \textbf{Right:} Final accuracy broken down by EC class.}
\label{fig:ecnumber_curves}
\end{figure*}

\begin{figure*}[h]
\centering
\includegraphics[width=\textwidth]{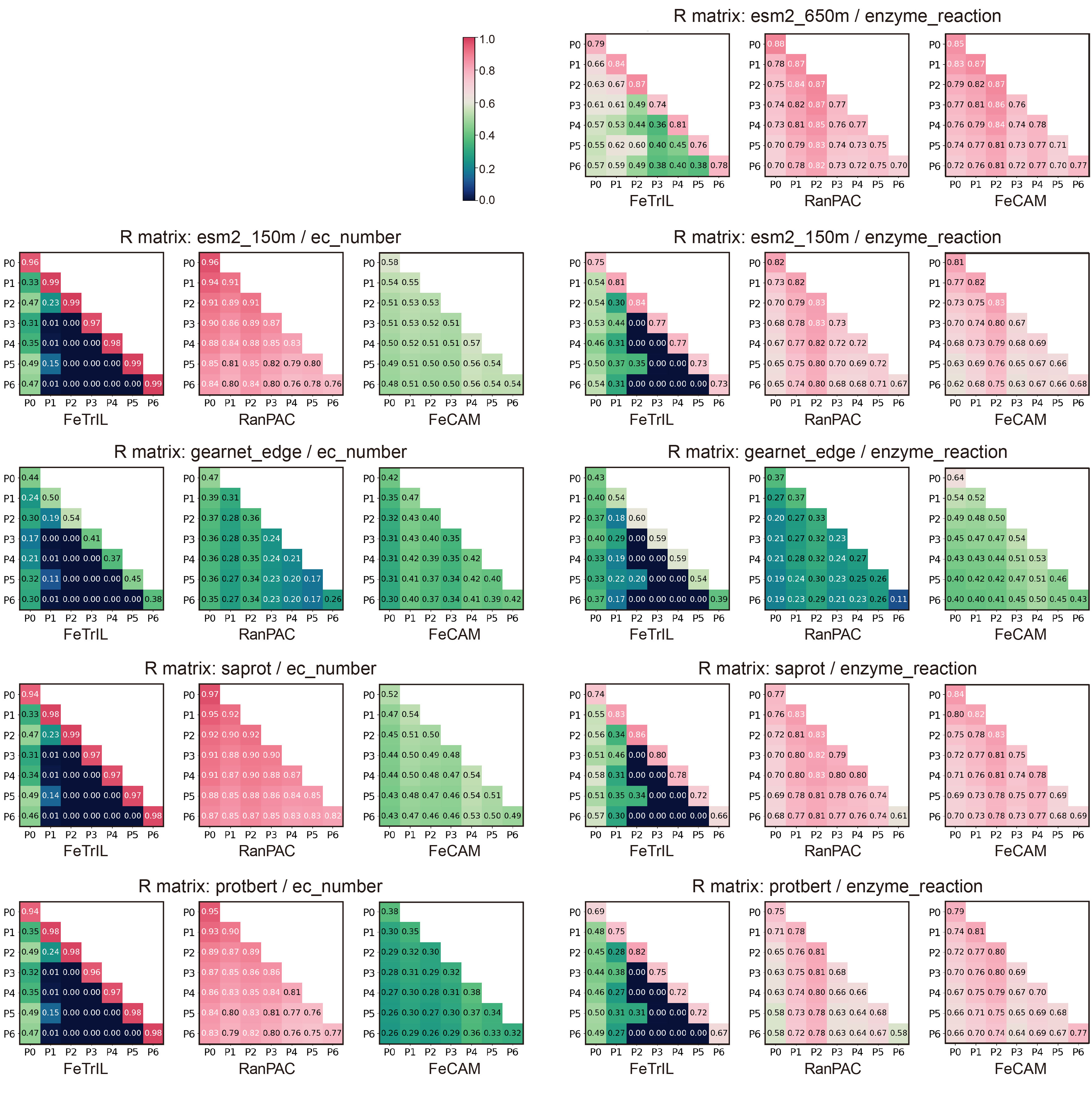}
\caption{Phase-wise accuracy matrices across encoder--dataset configurations. Within each configuration, matrices from left to right correspond to FeTrIL, RanPAC, and FeCAM. Rows indicate the training phase and columns the evaluated phase.}
\label{fig:appendix_rmatrix}
\end{figure*}


\end{document}